\def\year{2022}\relax
\documentclass[letterpaper]{article} 
\usepackage{aaai22}  
\usepackage{times}  
\usepackage{helvet}  
\usepackage{courier}  
\usepackage[hyphens]{url}  
\usepackage{graphicx} 
\usepackage{natbib}  
\usepackage{caption} 
\DeclareCaptionStyle{ruled}{labelfont=normalfont,labelsep=colon,strut=off} 
\usepackage{algorithm}
\usepackage{algorithmic}

\usepackage{epstopdf}

\graphicspath{{./images/}} 

\usepackage{graphicx}
\usepackage{amsmath,amssymb}
\usepackage{adjustbox}
\usepackage{mathtools}
\usepackage{booktabs}
\usepackage{xcolor}
\usepackage{cases}
\usepackage{subcaption}

\definecolor{citecolor}{HTML}{0071bc}
\usepackage[pagebackref,breaklinks=true,letterpaper=true,colorlinks,citecolor=citecolor,bookmarks=false]{hyperref}

\title{Fast and Data Efficient Reinforcement Learning from Pixels via Non-Parametric Value Approximation}

\author {
    Alexander Long,\textsuperscript{\rm 1}
    Alan Blair, \textsuperscript{\rm 1}
    Herke van Hoof \textsuperscript{\rm 2}
}
\affiliations {
    \textsuperscript{\rm 1} University of New South Wales\\
    \textsuperscript{\rm 2} University of Amsterdam\\
    alex.long@unsw.edu.au,  blair@cse.unsw.edu.au, h.c.vanhoof@uva.nl
}

\DeclareMathOperator*{\argmax}{arg\,max}
\newcommand{\normltwo}{L^2}

\begin{document}

\maketitle
	\begin{abstract}
	We present Nonparametric Approximation of Inter-Trace returns (NAIT), a Reinforcement Learning algorithm for discrete action, pixel-based environments that is both highly sample and computation efficient. NAIT is a lazy-learning approach with an update that is equivalent to episodic Monte-Carlo on episode completion, but that allows the stable incorporation of rewards while an episode is ongoing. We make use of a fixed domain-agnostic representation, simple distance based exploration and a proximity graph-based lookup to facilitate extremely fast execution. We empirically evaluate NAIT on both the 26 and 57 game variants of ATARI100k where, despite its simplicity, it achieves competitive performance in the online setting with greater than 100x speedup in wall-time.
	
\end{abstract}

\section{Introduction}
Deep Reinforcement Learning (DRL) has become the standard approach to tackle complex non-linear sequential decision making problems exhibiting a high number of observational variables~\cite{Li2017}, a subset of which are image, or pixel, based. Although tabular Reinforcement Learning (RL) algorithms will learn (and in fact provably converge) in such environments, they are too slow to be practically useful. Policy and value function parametrization is thus typically used to facilitate learning, which has led to several breakthrough results ~\cite{dota, badia2020agent57}.

\begin{figure}[t]
	\centering
		\includegraphics[width=0.9\columnwidth]{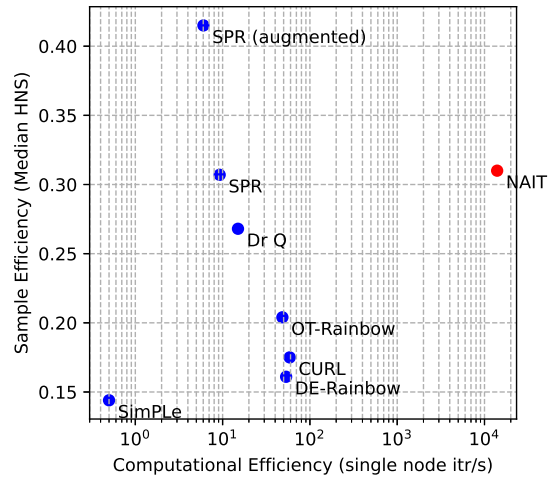}
		\label{fig:final}
	\caption{Comparison of our approach to other online sample efficient RL algorithms baselined on ATARI100k. NAIT achieves a $>$100x speedup over all other methods and is only outperformed in terms of median HNS by augmented SPR. Up and to the right is better. Algorithm labels are provided in Related Work.}
\end{figure}

Advances in raw performance, however, have been achieved at the cost of increased sample, and computational complexity. Agent 57~\cite{badia2020agent57} for example, requires 256 actors collecting approximately 57 years of real-time game play per game. For systems tackling more complex environments, these numbers are significantly higher. In an online,  real-world setting this presents a major roadblock and, as a consequence, research attention has begun to shift towards improving sample efficiency with the goal of broadening the applicability of Deep Reinforcement Learning (DRL) in real world control and robotics. In these domains, sample efficiency is critical as the cost of each interaction is high; a physical robot must actually move. Improvements in sample efficiency, however, typically come at the cost of increased computational complexity, where additional CPU cycles are used to ``squeeze'' more from the available data, through a learned model ~\cite{dreamer}, data augmentation ~\cite{RAD}, or similar. 

The dominant evaluation framework for online RL algorithms remains simulated environments. In this setting, high computational complexity is not a bottleneck on performance. It does, however, decrease the practical relevance of the associated results by increasing the cost (in dollar terms), and thus reducing accessibility~\cite{rlthatmatters}. Scientific progress in the area is also impacted, as is visible in the fact that the high computational cost of SimPLe~\cite{Kaiser2019} was one cause of the choice to evaluate on the 26 game ``easy exploration'' subset of the full 57 game suit, which made comparison to previous work difficult. 
It also make comparisons difficult, and expensive, for subsequent authors.

Computationally efficient approaches increase the scale at which algorithms can operate. For this reason several works have made attempts to speed up the evaluation simulators, such as CuLE for ATARI \cite{gpuatari}. Few, however, have focused on improving the computational efficiency of the RL algorithms themselves, instead focusing on distributed implementations \cite{acceleratedRL}, or the use of specialized hardware \cite{fpgaatari}. Increasing the wall-clock speed with which decisions can be reached also directly increases the responsiveness of the system, which greatly simplifies tasks such as tracking and control~\cite{Wagner5226627}.

In this paper, we present Nonparametric Approximation of Inter-Trace returns (NAIT), an algorithm that is \textit{both} computation and sample efficient. NAIT makes use of fast approximate k-nearest neighbour algorithms to generalise locally, and approximate a value function from stored memory. We combine this with a novel inter-trace update that provides the stability of episodic returns as well as the ability to immediately learn from an experience prior to completion of the episode (an advantage that is particularly pronounced in the low sample domain). We make use of a Discrete Cosine Transform (DCT) based transformation as the state representation, with no use of a learned encoder, prior knowledge, domain specific observation transformations or data augmentation.

We focus our evaluation on the ATARI100k suite of environments ~\cite{Kaiser2019} in the online setting. Our approach is able to achieve 0.31 Median Human Normalized Score (MHNS), outperforming several previous state-of-the-art approaches such as Dr. Q \cite{Kostrikov2020} and CURL \cite{curl}, despite making no use of a learned representations. In addition to this high level of sample efficiency, our approach is over two orders of magnitude faster than the current state of the art, SPR \cite{Schwarzer2020}, in terms of wall-time. In a single thread, NAIT is able to complete training for one environment in 7.2 minutes, where SPR requires 3.5 hours on the same GPU-enabled node. We examine our approach with ablation and sensitivity analysis, demonstrating the ability of NAIT to trade-off performance for speed, as well as a high level of robustness across hyperparameter values. Because of NAIT's low execution time, we are able to provide baseline results under multiple evaluation settings and sample budgets, providing a strong baseline for future work.

\section{Related Work}
Recently significant focus has turned towards sample efficient RL in image-based environments. SimPLe~\cite{Kaiser2019} consists of a novel network architecture used to model ATARI forward dynamics in the observation space, integrated within a Dyna-style~\cite{Sutton1990} training loop. This work introduced the ATARI100k benchmark, and achieved $0.144$ median human normalised score (MHNS), significantly outperforming Rainbow~\cite{Hessel2017} and PPO~\cite{Schulman2017}. Data Efficient Rainbow (DE-Rainbow) \cite{VanHasselt2019} was introduced to make use of more frequent value network updates per environment step, matching the update/step ratio used in SimPLe and achieved 0.161 MHNS while making no use of a learned model. Concurrently, Kielak presented OT-Rainbow~\cite{importanceof}, a similarly ``over-trained'' but otherwise unaltered version of rainbow capable of achieving $0.26$ MHNS.

Several works have identified representation learning as the bottleneck in this domain and have proposed techniques to accelerate it. CuRL~\cite{curl} introduced an auxiliary contrastive loss able to be utilized within existing RL algorithms. Dr. Q ~\cite{Kostrikov2020}, focused solely on image augmentation (random shifts and random crops of observations), and demonstrated their inclusion raised the performance of an efficient DQN implementation from $0.058$ to $0.284$ MHNS. SPR~\cite{Schwarzer2020} proposes a latent space forward dynamics model, combined with an auxiliary prediction loss, that achieves 0.415 MHNS when combined with an augmentation scheme similar to that in Dr. Q. 

Recently, several works such as APT \cite{APT} and VISR \cite{VISR} have considered an additional unsupervised pretraining phase prior to online learning. In this setting, agents are provided a 250M frame budget with no environment rewards, after which rewards are restored for the standard 100k steps. Although we do not adopt this setting to evaluate out agent, we include several results to help contextualize the performance of NAIT.

Non-parametric, or lazy learning ~\cite{Sutton1998} has been applied to ATARI previously. In the original work proposing the ALE environment~\cite{Bellemare2015}, a Locality Sensitive Hashing (LSH) algorithm was proposed to convert pixel observations to binary features for SARSA($\lambda$). Linear learners ingesting fixed features were examined in \cite{shallowatari}, and remain strong baselines. In Model Free Episodic Control (MFEC)~\cite{Blundell2016}, a gaussian random projection to a 64-dimensional latent space is used on ATARI frames, however performance is poor, possibly due to the small latent space dimensionality chosen. Berariu et. al.~\cite{Berariu2017} augmented MFEC with sparse projections~\cite{Achlioptas2003} along with variance normalisation in the projected space and demonstrated a improvement in performance, along with significant computational gains on ATARI.

\section{Method}
\subsection{Background}
We consider the standard formulation and notation~\cite{Sutton1998} of an agent interacting with some environment at discrete time steps to maximise total expected reward. The environment is modelled as a Partially Observable Markov Decision Process (POMDP)  $\left(\mathcal{O}, \mathcal{A}, P, r, \gamma\right)$ with a $D$ dimensional observation space $o\in \mathcal{O}: \mathbb{R}^D$ and discrete action space $a\in \mathcal{A}$. 
Traces are denoted as 
$
\tau = \{(o_0, a_0, r_0), (o_1, a_1, r_1), \ldots\}
$.
$P=\text{Prob}(o_t'|o_{\leq t}, a_{\leq t})$
denotes the transition probability dynamics, $r: \mathcal{O} \times \mathcal{A} \rightarrow \mathbb{R}$ denotes the reward function and $\gamma \in [0, 1)$ denotes a discount rate. We make use of a fixed projection $\phi: \mathbb{R}^{4\times D} \rightarrow \mathbb{R}^F$, to convert a stack of four observations (as is standard in ATARI) to a more compact space $s \in \mathcal{S} : \mathbb{R}^F$, $s_t = \phi\left(o_t, o_{t-1}, o_{t-2}, o_{t-3}\right)$ where $F\ll D$. We treat $s$ as the base state henceforth, with $P=\text{Prob}(s_t'|s_t, a_t)$ and $r_t=r(s_t, a_t)$.
The discounted return from state $s_t$ and action $a_t \in \mathcal{A}$ taken at time $t$, is defined as $G(s_t, a_t) \doteq \sum_{k=t}^{H} \gamma^{k-t} r(s_k, a_k)$ where $H$ is the time horizon of the episode. For brevity, we often abbreviate this as $G_t = G(s_t, a_t)$ indicating the return from time $t$ onwards. If a range is specified as in $G_{t:T}$, this denotes the return from time $t$ until $T$. Our goal is to learn the optimal policy $\pi^*(s_t)$ that maximises $\mathbb{E}_P\left[G(s_t, a_t)\right]$ for all $s_t \in \mathcal{S}$ where $a_t \sim \pi^*(s_t)$. We make use of a memory $M$, where $M_a(s)$ refers to the value in memory for action $a$ and state $s$.

\begin{algorithm*}[ht]
	\caption{NAIT: Training and Acting}
	\label{alg:full}
	\begin{algorithmic}
		\STATE {\bfseries Initialise:} 
		Memory $M_a \rightarrow \{\}$ for all $a\in\mathcal{A}$
		\WHILE{\textit{training}} 
		\STATE $o \leftarrow  $ reset environment, empty trace history $\tau \leftarrow \{\}$,  $\mathbf{G} \leftarrow \{\}$,  $\mathbf{\Gamma} \leftarrow \{\}$,  $M'_a \xleftarrow{\text{copy}} M_a$ for all $a$
		\FOR{$t=0$ to epsisode horizon $H$} 
		\STATE Get DCT representation $s \leftarrow \phi\left(o_t, o_{t-1}, o_{t-2}, o_{t-3}\right)$ 
		\FOR{$a \in \mathcal{A}$} 
		\STATE Get estimate of $Q(s, a)$ using a kernel $K(s, s')$, set of neighbours $L(s, a) = \{s_1,\dots, s_k\} $
		\STATE $\qquad 	Q(s, a;k)=
		\begin{cases}
		M_a(s) & s \in M_a \\
		\frac{\sum_{s'\in L(s, a)} K\left(s, s'\right) M_a(s')}{\sum_{s'\in L(s, a)} K\left(s, s'\right)}, & s \notin M_a, \; |M_a| \geq k  \\
		0 & s \notin M_a, \;  |M_a| < k
		\end{cases}$

		\ENDFOR

		\STATE $\pi(s) = \argmax_{a\in\mathcal{A}}Q(s, a)$ breaking ties with $\pi_{\text{tiebreak}}(s) \sim \text{softmax}_{a\in \text{C}}\left(\frac{1}{k}\sum_{s' \in L(s, a)}||s - s'||_2^2\right)$
		\STATE Act $o, r \leftarrow \text{env}(\pi(s))$
		\STATE Add $(s, a) $ to $\tau$ 
		\STATE Get inter-trace targets (Alg. \ref{alg:vecintertrace}) and  Update($\tau, \mathbf{Q}_{\textrm{IT}}$) (Alg. \ref{alg:update})
		\ENDFOR
		\STATE Get final returns and Update($\tau, \mathbf{G}$) (Alg. \ref{alg:update}) 
		\ENDWHILE
	\end{algorithmic}
\end{algorithm*}

At a high level, our approach acts by representing all observations via a fixed projection, calculating Q-values via local approximation, and updating these values iteratively as new rewards are encountered. The final algorithm is shown in Algorithm \ref{alg:full}. We outline each component below.

\subsection{Representation}
As a fixed representation, we use a Discrete Cosine Transform (DCT) and select the first $F$ DCT coefficients ordered by frequency. The major advantage of the DCT as representation is that it is able to capture broad detail about the image extremely compactly, while being fast to compute, as $\mathcal{O}(D\log D)$ implementations based on the FFT exist.  This is faster even than a random projection at $\mathcal{O}(FD)$ for many values of $F$. We make use of DCT II, an orthonormal transform $\mathbb{R}^{N} \rightarrow \mathbb{R}^{N}$,  
\begin{equation}
\text{DCT:} \, X_{i}=\sum_{n=0}^{N-1} x_{n} \cos \left[\frac{\pi}{N}\left(n+\frac{1}{2}\right) i\right]  i=0, \ldots, N-1.
\end{equation}
We wish to apply this to the frame-stack obtained from the environment, and thus stack the four frames in a 2x2 grid to ensure equal weighting of vertical and horizontal frequencies, and apply the 2D DCT II. To compress the state, we discard high frequency coefficients and select the top $F$;
\begin{align}
s_t &= \phi\left(o_t, o_{t-1}, o_{t-2}, o_{t-3}\right)\\\nonumber
&= \left[ \text{DCT}\left(\text{DCT}
\begin{bmatrix}
o_t & o_{t-1}\\
o_{t-2}& o_{t-3}\\
\end{bmatrix}
^T\right)^T_{ij}\right]_{0 \le i, j \le \sqrt{F}}.
\end{align}

As the DCT is the only component of our approach specific to images, we also present results with domain-agnostic random projections as the representation (see Table \ref{tab:ablations}). A random projection transforms the observation data (of dimension $D$) to a lower dimensional space (of dimension $F$) $\phi : x \rightarrow x\textbf{R}$, where $\textbf{R} \in \mathbb{R}^{D \times F}$. We fix $F$ to be the same as the number of DCT coefficients used.
For each element $u_{i, j} \in \textbf{R}$:
\begin{equation}\label{eq:projection}
u_{i, j}=\sqrt{s}
\begin{cases}
+1 & \text { with prob. } \frac{1}{2 s} \\ 
0 & \text { with prob. } 1-\frac{1}{s} \\
-1 & \text { with prob. } \frac{1}{2 s} 
\end{cases}.
\end{equation}

When $s = \sqrt{D}$, $\textbf{R}$ is a \textit{very sparse random projection}~\cite{Li2006}. When $s=3$, the projection becomes the more standard \textit{sparse random projection}~\cite{Achlioptas2003}.

\subsection{Non Parametric Value Function Approximation}
We aim to learn a local approximation of the the discounted return $Q(s, a) \simeq G(s, a)$ where $a \sim \pi ^*(s)$. $Q$-values are estimated from action-specific memory buffers $M_a$, of which frozen copies, $M'_a$ are made at the start of each episode,  $M'_a(s) \xleftarrow{\text{copy}} M_a(s)$ at $t=0$. This memory is updated online with learning rate $\alpha$ according to the learning rule in Alg. \ref{alg:update}. Q-values are directly estimated from memory,  
\begin{equation}\label{eqn:estimate}
Q(s, a |k)=
\begin{cases}
M_a(s) & s \in M_a \\
\frac{\sum_{s'\in L(s, a)} K\left(s, s'\right) M_a(s')}{\sum_{s'\in L(s, a)} K\left(s, s'\right)},\!\!\!\!   & s \notin M_a,  |M_a| \geq k  \\
0 & s \notin M_a,   |M_a| < k
\end{cases}
\end{equation}
where $L(s, a) = \{s_1,\dots, s_k\} $ is the result of a nearest-neighbour (by $L^2$ distance) lookup of $s$ on  $M_a$, and $K: \mathcal{S}\times\mathcal{S} \to \mathbb{R}$ is a kernel function that dictates a weighting over neighbours. In all experiments, to smooth the interpolation, we use a normalized tricubic kernel. The kernel calculates the weighting for each neighbour $s' \in L(s, a)$ as,
\begin{equation}
K(s, s') = \left( 1- \left(\frac{||s - s'||_2}{\max_{\hat{s}\in L(s,a)}||s - \hat{s}||_2}\right)^3 \right)^3.
\end{equation}

\begin{algorithm}[ht]
	\caption{Batch Update}
	\label{alg:update}
	\begin{algorithmic}
	\STATE \textbf{Input} $\tau$: $(s, a)$ tuples list, $\mathbf{Q}$: targets list, $\alpha$: learning rate
	\FOR{$(s, a), Q \in \textrm{zip}(\tau, \mathbf{Q})$}
    	\IF{$s \in M'_a$}
    	\STATE $M_a(s) \leftarrow M'_a(s) + \alpha \left(Q -M'_a(s) \right)$
    	\ELSE
    	\STATE $M_a(s) \leftarrow Q$
    	\ENDIF
    \ENDFOR
\end{algorithmic}
\end{algorithm}

Normalization is required as the standard kernel has support $[0, 1]$ and the distance between two states can clearly exceed this. Consequently, we normalize by the maximum $\normltwo$ distance to all states in $L(s, a)$. This has significant consequences in that the bandwidth of the kernel expands and contracts depending on the density of the local neighbourhood at the query point due to the fixed $k$. This results in broad generalisation across large areas when samples are sparse, and fine-grained interpolation when a region has been heavily sampled. The particular choice of kernel is not a critical component of NAIT and a gaussian, bicubic or similar may easily be substituted. Empirically, we found the tricubic kernel to give a slight bump in performance (see Table \ref{tab:ablations}).

Acting is done greedily with ties broken by an estimate of the visitation density calculated with information from the already completed $k$-nn lookup. This is done by a softmax distribution over the mean $\normltwo$ distance to the neighbouring states found on the lookup for that $(s, a)$ pair (if there is an exact match, the distance is considered 0). Given a candidate set of maximum valued actions $C = \argmax_{a\in\mathcal{A}}Q(s, a)$, 
\begin{equation}\label{eq:policy}
\pi_{\text{tiebreak}}(s) \sim \text{softmax}_{a\in \text{C}}\left(\frac{1}{k}\sum_{s' \in L(s, a)}||s - s'||_2^2\right).
\end{equation}
If there is no tiebreak ($|C|=1$), the policy is standard greedy action selection, $ \pi \sim \argmax_{a\in\mathcal{A}}Q(s, a)$. This distance-based tie-breaking is the only form of exploration explicitly added to NAIT.

\subsection{Inter-Trace Update}\label{sec:inter} 
Monte-Carlo returns are desirable in the low-sample domain as they do not propagate approximation error. However, they come with the major drawback that updates cannot be carried out until the end of a trace. When the ratio of average trace length to total interactions is high, as it is in ATARI100k, this property has a pronounced effect. To address this, we introduce a novel inter-trace update allows the incorporation of new information immediately upon receiving it, as is possible with bootstrapped updates, while retaining the stability of MC returns. We show this update can be calculated in constant time with respect the episode length.

During an ongoing trace, at time $t<T$, for an $(s, a)$ pair first experienced at time $i$, the standard MC target $Q_{\text{target}}(s, a) = G_{i:T}(s, a)$, is not able to be calculated as the episode has not completed. We can, however, incorporate the rewards $\left\{ r_i \dots r_t\right\}$ already received into a new target as an approximation to the MC return, bootstrapping off a fixed copy, $Q'(s, a)$ of the current estimator. Here $Q'(s, a)$ indicates Eq. \ref{eqn:estimate}, calculated with the frozen memory $M'_a$. This new target is;
\begin{align}\label{eq:interim_target}
\begin{split}
Q_{\text{target}}(s_i, a_i) & = r_i + \gamma r_{i+1} + ... + \gamma^{t-i}r_t  \\
& \qquad + \gamma^{t-i+1}Q'(s_{t+1}, a_{t+1})\\
&= G_{i:t-1} + \gamma^{t -i}r_t + \gamma^{t-i+1}{Q'(s_{i+1}, a_{i+1})}
\end{split}	
\end{align}

In practice we vectorize this operation so as to calculate the targets for all $(s,a)$ pairs visited in the trace in one step, ensuring updates can be computed in constant time. Specifically, at each time $t$ during a trace, we have encountered $n$ unique $(s, a)$ pairs, with the time of their first visit given by $\mathbf{t} = [t_0, t_1, ... , t_n]$. We can compute the target vector,
\begin{align}
\mathbf{Q}_{\textrm{target}} &=[Q_{\textrm{target}}(s_{t_0}, a_{t_0}), ..., Q_{\textrm{target}}(s_{t_n}, a_{t_n})]\\
&= \mathbf{G} + \mathbf{\Gamma}Q'(s_{t+1}, a_{t+1})
\end{align}
where 
\begin{align}
\mathbf{G} &= [G_{t_0:t}, G_{t_1:t}, ..., G_{t_n:t}],\\
\mathbf{\Gamma} &= [\gamma^{t-t_0}, \gamma^{t-t_1}, ..., \gamma^{t-t_n}] 
\end{align}
$\mathbf{G}$ is simply the running discounted reward, and $\mathbf{\Gamma}$ the discount that should be applied to the next reward, for each $(s, a)$. At the beginning of a trace, all vectors are cleared, then are iteratively updated at each time-step following Alg. \ref{alg:vecintertrace}.

\begin{algorithm}[ht]
	\caption{Vectorized Inter-trace Targets}
	\label{alg:vecintertrace}
	\begin{algorithmic}
	\STATE \textbf{Input} $r_t$, $Q'(s_{t+1}, a_{t+1})$,  $\mathbf{G}, \mathbf{\Gamma}$
	\STATE Append 1 to $\mathbf{\Gamma}$
	\STATE $\mathbf{G} \leftarrow \textbf{G} + \mathbf{\Gamma}r_t$
		\STATE $\mathbf{\Gamma} \leftarrow \gamma \mathbf{\Gamma}$
	\STATE $\mathbf{Q}_{\textrm{IT}} \leftarrow \mathbf{G} + \mathbf{\Gamma}Q'(s_{t+1}, a_{t+1})$
	\STATE \textbf{Return } $\mathbf{Q}_{\textrm{IT}}, \mathbf{G}, \mathbf{\Gamma}$
	\end{algorithmic}
\end{algorithm}

This update allows us to perform training once per step, as opposed to once per episode, instantly incorporating new information into our policy. The inter-trace update is similar to an \textit{n}-step update, but $n$ is always the maximum possible value for each state, and can be thought of as ``doing as much MC as possible'' - overwriting previous updates as new information becomes available, as opposed to bootstrapping off them. This is an important property as it does not propagate errors present in the value function approximation whereas standard \textit{n}-step updates do. While the calculation of targets can be done in constant time, each value update is $\mathcal{O}(1)$, and we do not need to update each pair at each step. Instead NAIT only utilizes new targets when $(t_0-t) \pmod r = 0$. In all experiments we set $r=50$. 

\subsection{Projected Small World Graphs}
Our method relies on being able to both add high-dimensional data to memory (Alg. \ref{alg:update}), and query this memory (eq. \ref{eqn:estimate}) for neighbours with low computational cost. With standard fast $k$-NN algorithms, such as KD-trees~\cite{kdtree}, this is not possible as they require re-indexing each time underlying data is changed which carries a large cost.

Instead, we make use of a Proximity Graph (PG) approach, Hierarchical Navigable Small World graphs (HNSW)~\cite{Malkov2016}. PG approaches are based around incremental index construction, and thus adding new points can be done cheaply. We make use of a specific standalone HNSW implementation~\cite{hnswlib} that leverages this feature to allow incremental updating, along with the ability to control search time parameters \texttt{ef} and \texttt{M} which directly effect the recall/query time trade-off (for full details see \cite{Malkov2016}). In addition, querying is an order of magnitude faster than a KD-tree on densely clustered data at high ($>0.8$) recalls~\cite{annbenchmarks}.

\section{Results}
\subsection{Image Random Walk}

\begin{figure}[ht]
\centering
	\begin{subfigure}[r]{0.21\textwidth}
		\includegraphics[width=\textwidth]{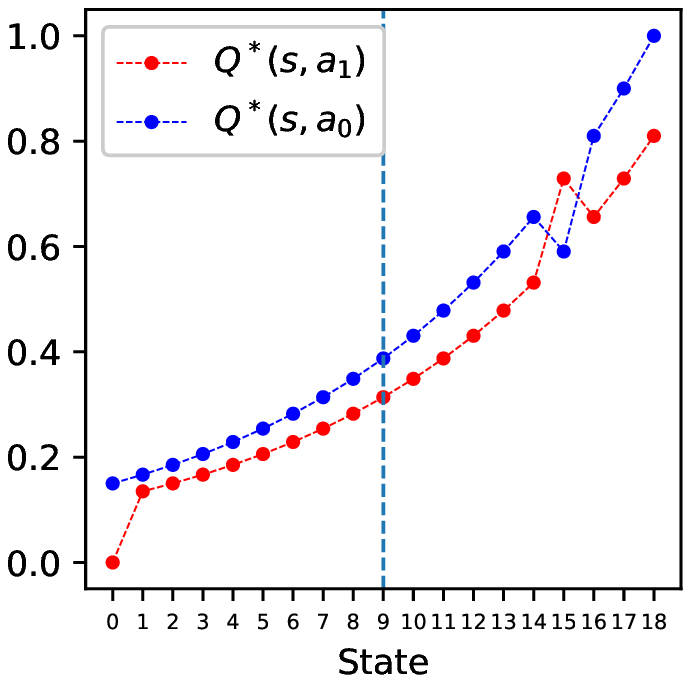}
		\caption{}
		\label{fig:optimalqs}
	\end{subfigure}
	\begin{subfigure}[r]{0.25\textwidth}
		\includegraphics[width=\textwidth]{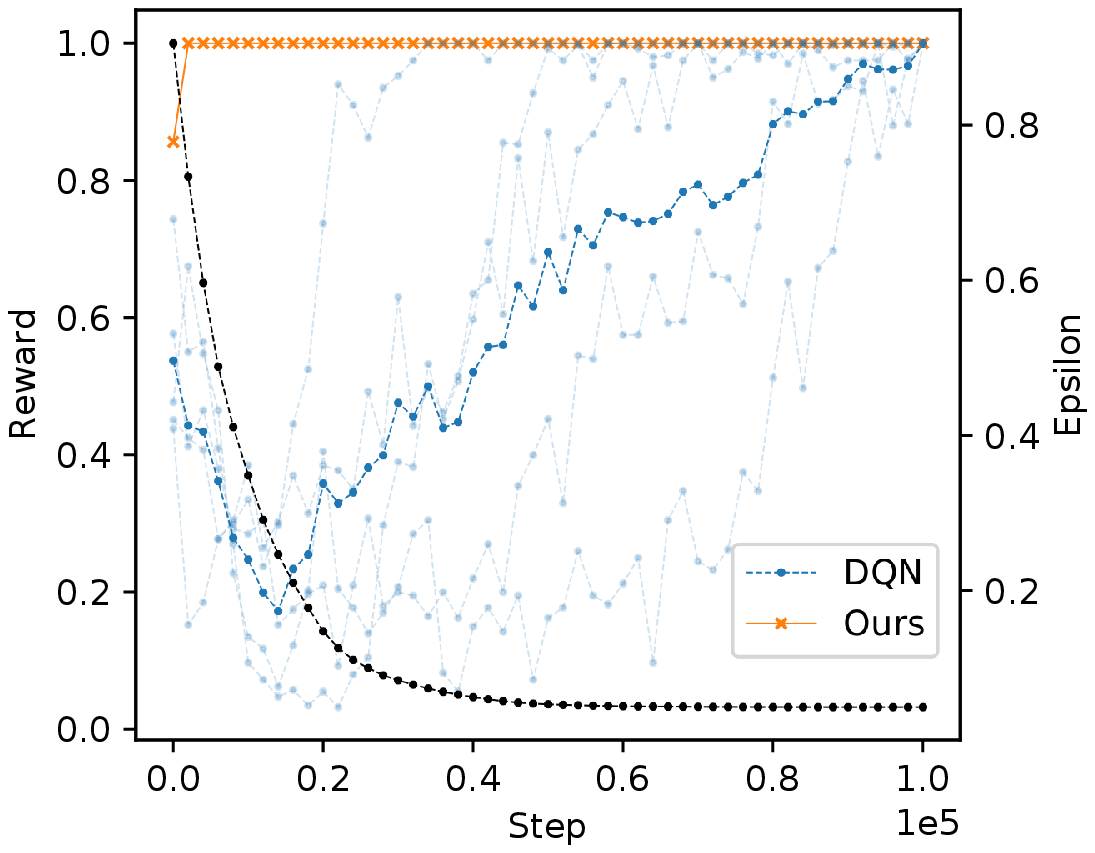}
		\caption{}
		\label{fig:RWreward}
	\end{subfigure}
	\caption{(a) Optimal Q-values for the Image Random Walk MDP (b) Performance on Image Random Walk over time. Exploration $\epsilon$ shown as black dotted line with scale on right axis. }
\end{figure}

As a base environment, we first introduce a 19-state deterministic MDP with image observations and $\gamma=0.9$. The 19 states are arranged in a flat line, in order of index. At each state two actions are available, $a_0$ to the left, action $a_1$ to the right. State 15 is special, here the actions are reversed (i.e. $a_0$ moves to the right, action $a_1$ to the left). A transition diagram is included in the supplementary material (Fig. \ref{fig:mdp}).

At all transitions the reward is zero, except for action $a_0$ from state 19, which yields a reward of 1. State 9 is the initial state. Observations for each state are noisy grayscale blocks, with the intensity corresponding to the states index $o_{i, j} = \mathcal{N}(n, \sigma), i, j \in \{1..D\}, n\in \{1..19\}$ where D is the height and width of the observations and $n$ is the index of the state, normalized to a range of $[-1, 1]$. We choose $D=5$ and $\sigma=0.1$. The optimal Q-functions for the system are shown in Fig \ref{fig:optimalqs}, where it can be seen that simple policies such as ``always move right'' cannot succeed due to the action switching at State 15. 

We also plot the reward (Fig \ref{fig:RWreward}) obtained at the end of each episode for a tuned DQN agent and our approach. Both methods settle on the optimal policy, however our memory based approach learns almost immediately, whereas the DQN is still converging after 100k steps. This result highlights that in some cases it is simply much easier to store visited returns and generalise locally at inference time, as opposed to trying to fit a global value function.

\subsection{ATARI100k}
\begin{figure}
\centering
	\includegraphics[width=0.83\columnwidth]{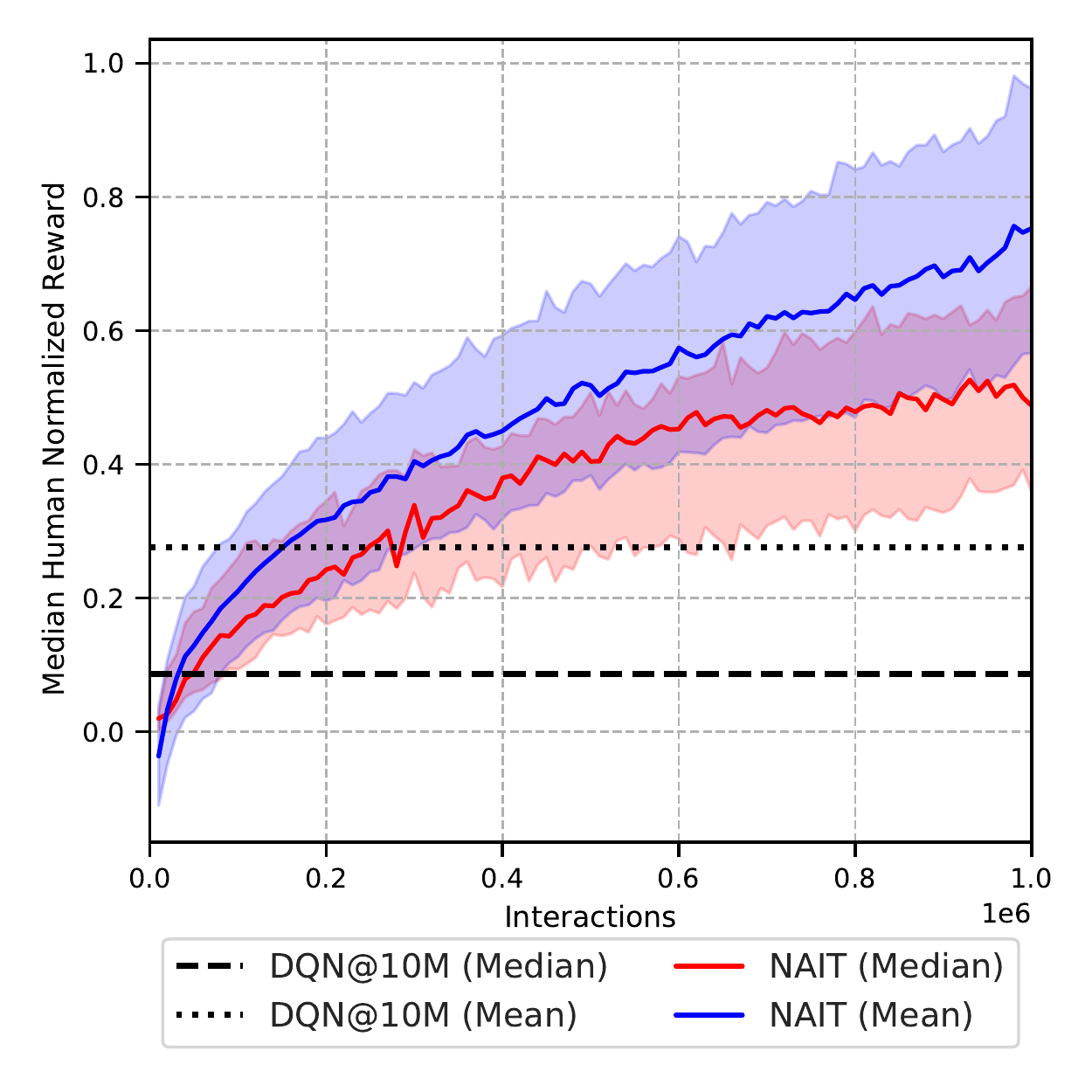}
	\caption{HNS over 1M interactions averaged over the full 57 ATARI game set. Shaded region indicates 90\% confidence interval. DQN performance after 10M interactions is also shown, as comparable sample efficient methods do not evaluate on the full game set. NAIT requires only 150k interactions to outperform DQN@10M on both metrics.}
	\label{fig:longrun}
\end{figure}

\begin{table}[t]
\begin{tabular}{llrrrr}
\toprule
                   &                  & \multicolumn{2}{c}{\textbf{ATARI100K}} & \multicolumn{2}{c}{\textbf{All 57 Games}}    \\ \midrule
\textbf{Method} & \textbf{Samples} & \textbf{Mdn}                                     & \textbf{Mean}                                   & \textbf{Mdn}          & \textbf{Mean}         \\ \midrule
\multicolumn{6}{c}{Online (low budget)}                                                                                                                                          \\ \midrule
SimPLe             & 100K             & 14.39                                            & 44.3                                            & -                     & -                     \\
DE-R               & 100K             & 28.5                                             & 16.1                                            & -                     & -                     \\
OT-R               & 100K             & 26.4                                             & 20.4                                            & -                     & -                     \\
CURL               & 100K             & 17.5                                             & 38.1                                            & -                     & -                     \\
DrQ                & 100K             & 28.42                                            & 35.7                                            & -                     & -                     \\
SPR$^\dagger$                & 100K             & \textbf{41.5}                                    & \textbf{70.4}                                   & -                     & -                     \\
NAIT               & 100K             & 31.21                                            & 36.23                                           & \textbf{17.57}        & \textbf{36.51}        \\ \midrule
\multicolumn{6}{c}{Online (medium budget)}                                                                                                                                        \\ \midrule
NAIT               & 250K             & 39.46                                            & 52.86                                           & 26.54                 & 48.81                 \\
PPO                & 500K             & 20.93                                            & 43.74                                           & -                     & -                     \\
NAIT               & 500K             & 46.73                                            & 69.97                                           & 43.63                 & 66.46                 \\
NAIT               & 750K             & 60.64                                            & 79.38                                           & 50.05                 & 73.26                 \\
NAIT               & 1M               & \textbf{60.60}                                   & \textbf{90.38}                                  & \textbf{52.04}        & \textbf{82.81}        \\
SARSA$^\ddagger$ & 2M & 35.85 & 72.40 & 28.27 & 57.75 \\
DQN                & 2M              & 27.8                                             & 52.95                                           & 8.61                  & 27.55                 \\ \midrule
\multicolumn{6}{c}{Unsup. Pretraining (250M budget) + Online Finetuning}                                                                                                \\ \midrule
VISR               & 100K             & 9.5                                              & 128.07                                          & 6.81                  & 102.31                \\
APT                & 100K             & 47.5                                             & 69.55                                           & \textbf{33.41}        & \textbf{47.73}        \\
APS$^*$            & 100K             & \textbf{51.45}                                   & \textbf{87.59}                                  & \textbf{-} \\ \bottomrule
\end{tabular}
\caption{Final performance measures. $^*$No shared encoder. $^\dagger$Data augmented. $^\ddagger$ SARSA($\lambda$) with Blob-PROST features.}
\label{tbl:main}
\end{table}

ATARI100k was first introduced as a testbed for SimPLe~\cite{Kaiser2019} and consists of 26 ``easy exploration'' ATARI2600 games. Since its introduction numerous works have focused solely on ATARI100k \cite{VanHasselt2019,importanceof,Kostrikov2020,curl,Schwarzer2020}. In our work, we report results for ATARI100k, but also evaluate on the 57 game set used more frequently in the wider RL literature. We do this both to provide a more robust baseline for future work, and to evaluate the generality of NAIT, as we do all hyperparameter tuning on the 26 game set alone, and make no changes when applying to the full evaluation suite. This setup was originally suggested by the ALE authors \cite{Machado2018}, and demonstrates that an approach is not overfitted to the task suit, however to date it has not been commonly adopted. We include a comparison to state-of-the-art sample efficient methods in Table \ref{tbl:main}. The unsupervised setting is included to contextualize NAIT's high level of performance on the full 57 game set, however we do not evaluate NAIT in this setting. Full settings for ATARI are listed in the supplementary material.

We run all experiments with 10 seeds, and evaluation is done on a rolling basis following the recommendations of ~\cite{Machado2018}, with the mean of the most recent 5 episodes logged for each seed. The DQN and SARSA($\lambda$) baselines are also reproduced from this work. Final scores are the mean score of 50 episodes, run after the completion of training, following~\cite{Schwarzer2020}. Our main metric is the Human Normalized Score, $\textrm{HNS} = \frac{\text{agent} - \textrm{random}}{\text{human} - \textrm{random}}$ achieved within a fixed interaction budget. We primarily report the Median HNS (MHNS) across games as the mean tends to be dominated by outlier performance.

As speed is a major feature of NAIT, we log median iterations per second (itr/s) across games. Itr/s is calculated  by dividing the time to complete the environment by 100k, averaged over ten seeds. Per-node speeds are the total throughput when all threads are allocated on a single AWS C5d.24xlarge (48 cores). This metric is indicative of real-world performance but is dependent on the number of cores present in the node.  For this reason we also report per-thread speeds which are independent of core count but do not consider factors such as I/O blocking, although in practice we observe these effects to be minor. SPR, SimPLE and DE-Rainbow speeds in Table \ref{tbl:speed} are reproduced from \cite{Schwarzer2020}, with Dr Q and CURL and OT-Rainbow bench-marked from from open-source implementations running on P3.16xlarge (32 core, 8xP100 GPU) instances.  

NAIT achieves a high level of performance, outperforming all methods with the exception of augmented SPR on the ATARI100K benchmark (see Table \ref{tbl:main}). This is achieved with a hugely reduced computational cost - over 100x faster than SPR and 1000x faster than SimPLe (see Table \ref{tbl:speed}). 

\begin{table}
\begin{tabular}{@{}lrr@{}}
\toprule
\textbf{Algorithm} & \multicolumn{1}{l}{\textbf{1 Node Itr/s}} & \multicolumn{1}{l}{\textbf{Walltime (hours)}} \\ \midrule
SimPLe             & 0.5                                       & 3166.67                                                        \\
DE-R               & 53.4                                      & 29.65                                                          \\
OT-R               & 48.4                                      & 32.71                                                          \\
CURL               & 58.4                                      & 27.11                                                          \\
DrQ                & 15                                        & 105.56                                                         \\
SPR                & 9.25                                       &   171.17                                                      \\
SPR (augmented)    & 6.03                                     &    262.58                                                        \\
NAIT       & \textbf{13962.24}                                  & \textbf{0.11}                                                           \\ \bottomrule
\end{tabular}
\caption{Average single-node iterations/s and walltime required to complete the full 57 game for various methods on a single node.}
\label{tbl:speed}
\end{table}

\begin{figure*}[t]
	\begin{subfigure}{0.24\textwidth}
		\includegraphics[width=\textwidth]{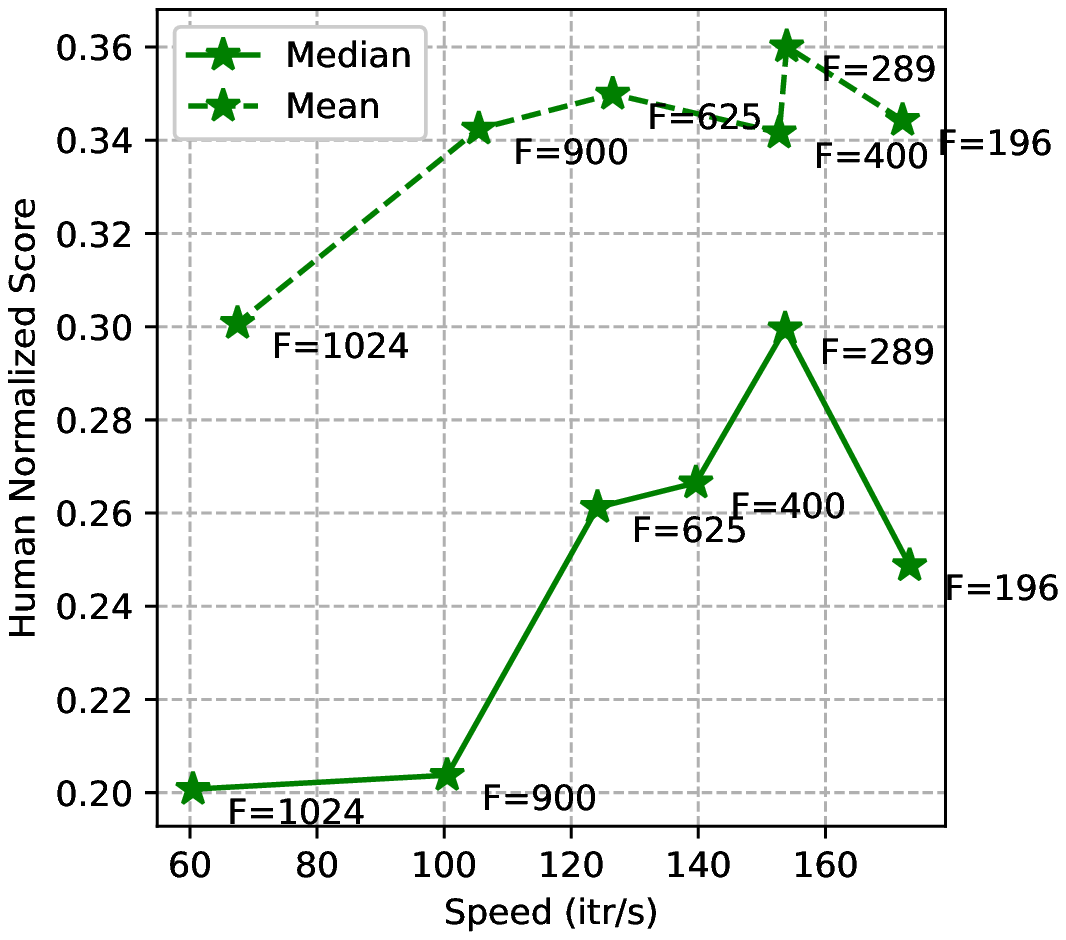}
		\caption{}
		\label{fig:k_sweep}
	\end{subfigure}
	\hspace{-10pt}
	\begin{subfigure}{0.24\textwidth}
		\includegraphics[width=\textwidth]{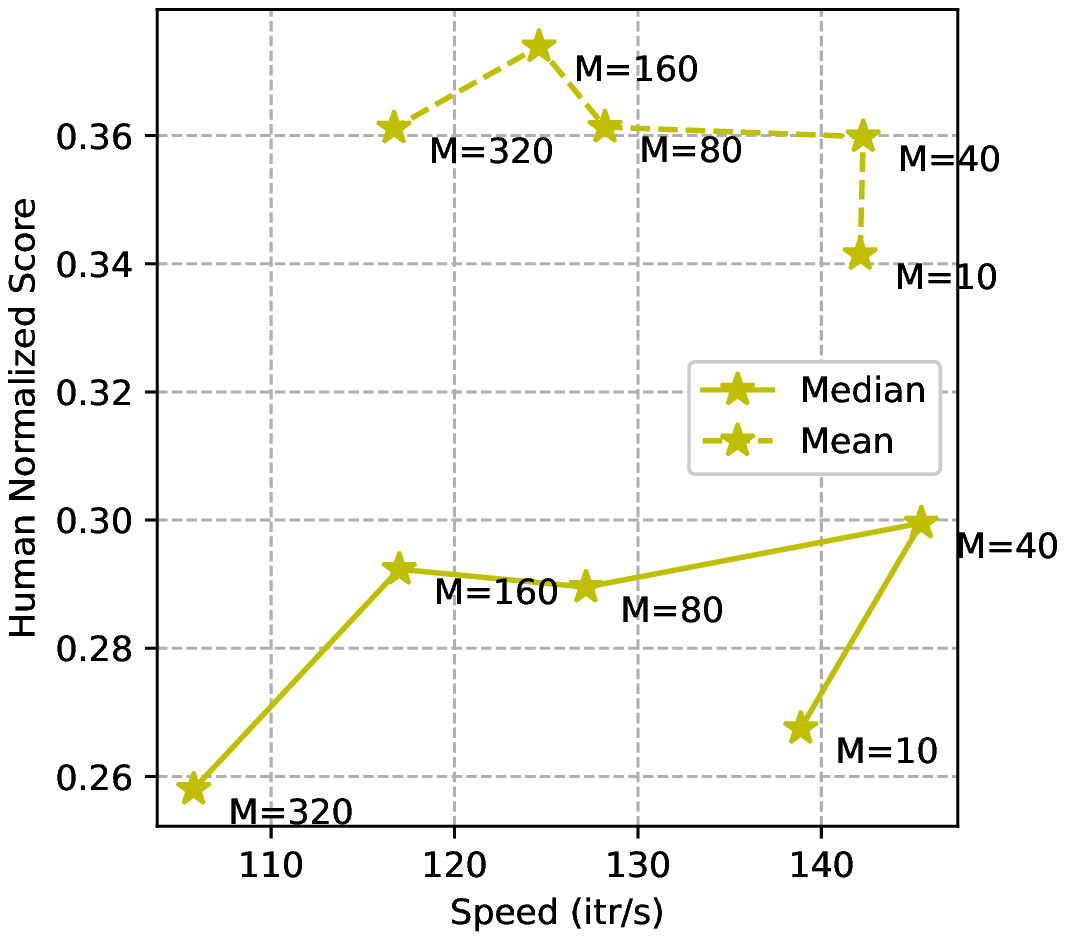}
		\caption{}
		\label{fig:n_sweep}
	\end{subfigure}
		\begin{subfigure}{0.26\textwidth}
			\includegraphics[width=\columnwidth]{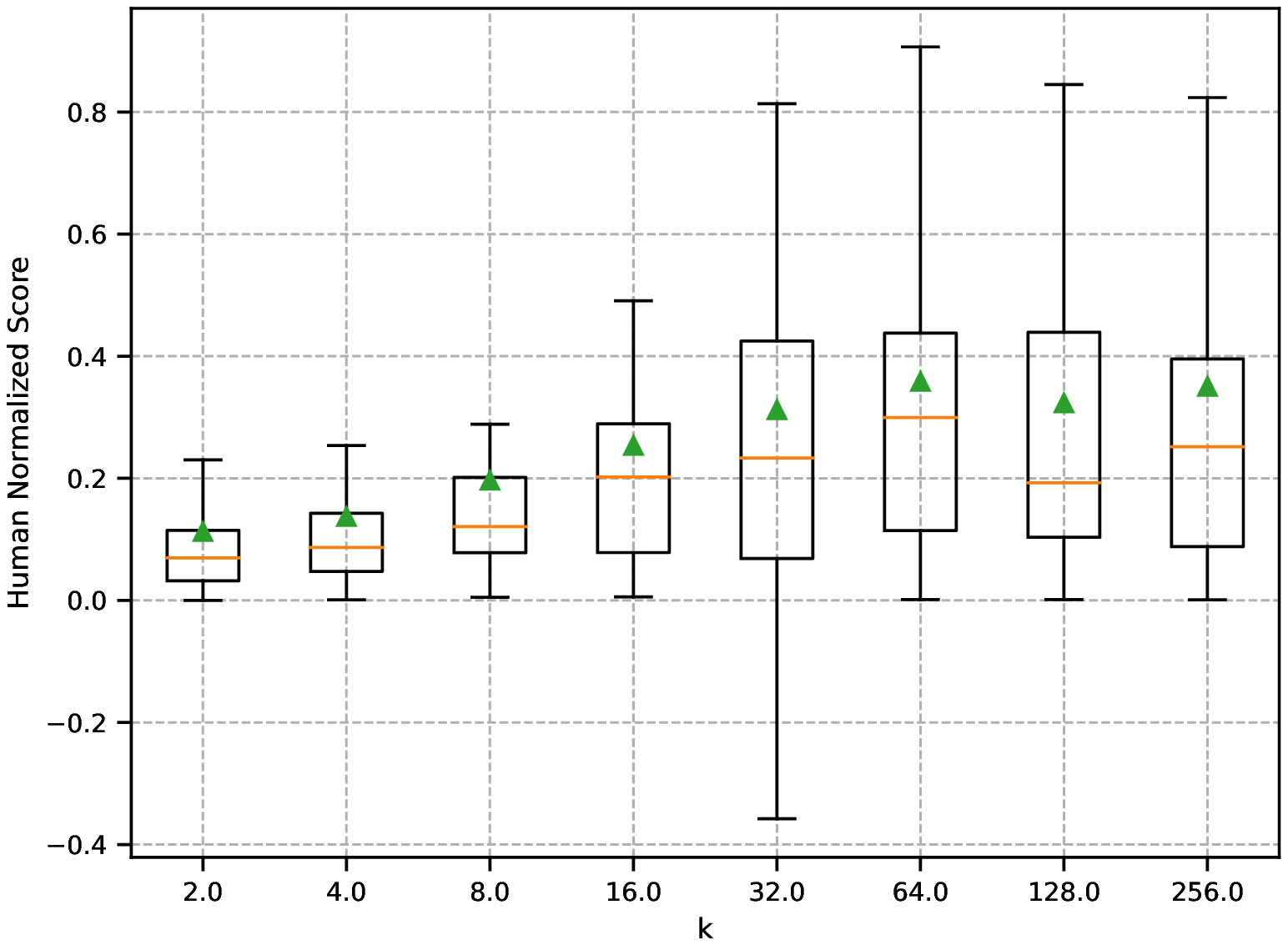}
			\caption{} 
			\label{fig:k_sweep_dists}
	\end{subfigure}
	\begin{subfigure}{0.26\textwidth}
			\includegraphics[width=\columnwidth]{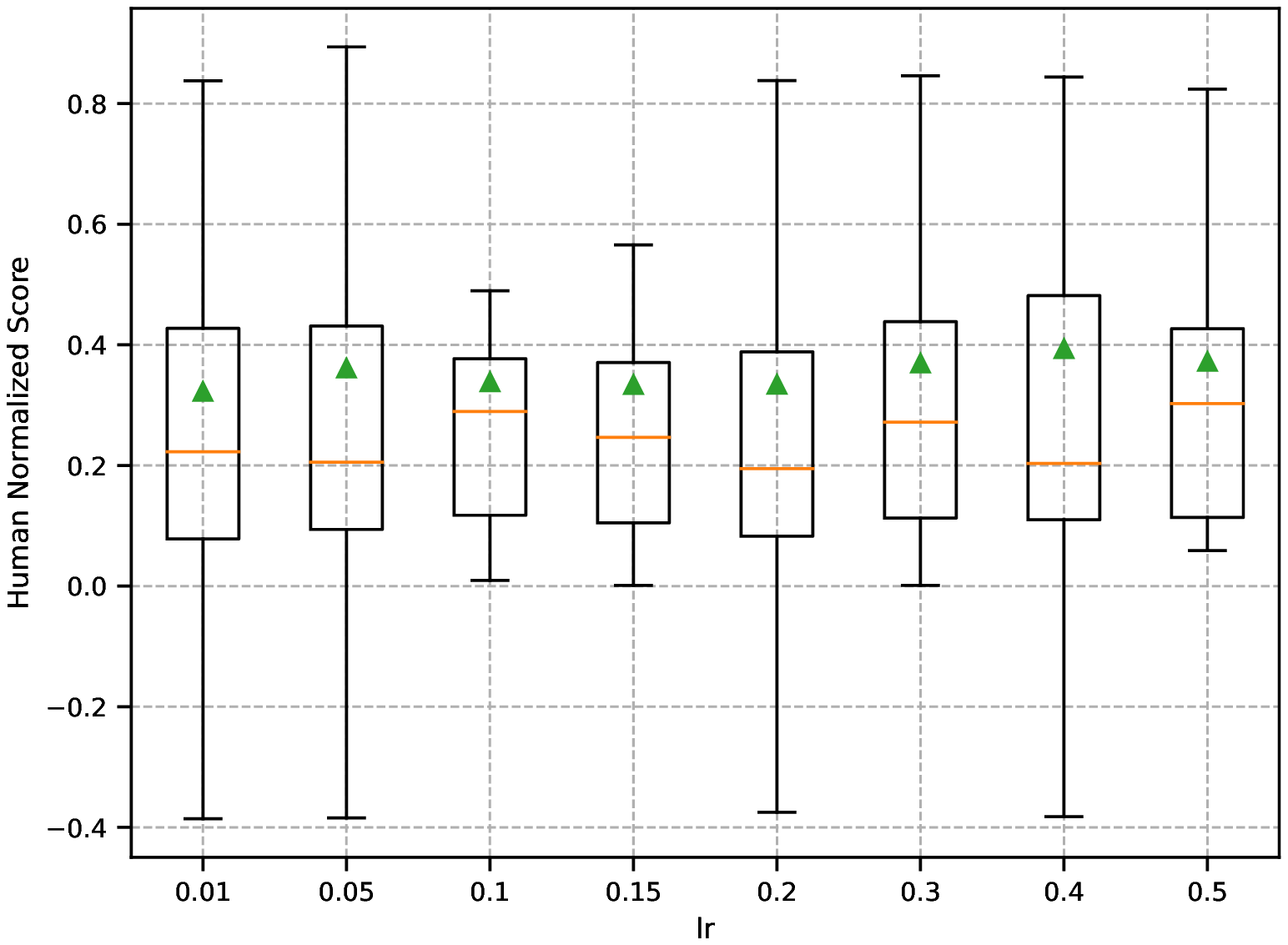}
			\caption{}
			\label{fig:n_sweep_dists}
	\end{subfigure}
	\caption{Sensitivity sweeps illustrating the sample efficiency (MHNS) and computational complexity (itr/s) trade-off by the choice of  (a) projected dimension $F$ (b) $M$, which affects the speed/recall ratio of HNSW. Up and to the right is better. For parameters that do not affect speed, we plot the distribution of final HNS's for c) $k$, and d) $\alpha$. Orange line indicates median, green diamond the mean. All game scores are first averaged over 10 seeds.}
	\label{fig:sweeps}
\end{figure*}

\section{Analysis}

\subsection{Ongoing learning}
NAIT demonstrates the high level of performance that a nonparametric approach can learn when environment interactions are limited. However, due to the fixed representation, NAIT's performance will saturate once the importance of a good representation (i.e. non-local generalization) outweighs the importance of quickly updating the value function or policy. As stated, the 100k domain is of interest in its own right, however we are also interested in quantifying the point at which design choices such as a fixed representation begin to limit performance. 

Using the same set of hyperparameters tuned for the ATARI100k benchmark (26 games, 100k interactions), we carried out a longer run over 1M interactions, and included all 57 games in the standard evaluation set (Fig. \ref{fig:longrun}). We made no other alterations to NAIT or its hyperparameters. Learning is highly stable and indicates NAIT possesses a high level generality. Per-game performance during training is visualized in the supplementary material.

\subsection{Ablations}\label{sec:ablation}

\begin{table}
	\begin{tabular}{llrr}
		\toprule
		\textbf{Component} & \textbf{Replacement} &\textbf{MHNS} &\textbf{Itr/s}\\
		\midrule
		NAIT& &\textbf{31.2} & 163.4\\		
		\midrule
		DCT projection& Very sparse & 21.5   & 121.3 \\
		DCT projection& Sparse & 28.1   & 91.4 \\
		DCT projection& None & - &  1.5   \\
		
		\midrule
		Tiebreaking & Uniform & 12.1 & 125.3 \\
		Intertrace update& Monte-Carlo & 13.6 &  141.2 \\
		Tricubic kernel&  Mean & 28.4 & 161.4 \\
		HNSW & KD tree & - & 0.3	\\
		\bottomrule
	\end{tabular}
	\caption{Ablations. Median HNS and single-thread itr/s reported for the 26 game subset and 100K iterations.}
	\label{tab:ablations}
\end{table}

When removing components, if a replacement is required, we choose what is either simplest or is already standard. For example, in place of a tricubic kernel, we use a uniform kernel, or instead of HNSW we use a KD-tree. Results are listed in Table \ref{tab:ablations}.

We find that DCT components provide better quality representations with regard to performance \textit{and} are faster to compute. Computation on the raw frames was prohibitively slow due to the resulting high dimensionality of the $k$-NN keys.

One unusual feature is the impact of distance based tiebreaking, in comparison to randomly breaking ties. This is likely because early in learning this encourages exploration based on novel states, as all estimates are zero, rather than simply ensuring a roughly uniform selection of actions. This is supported by the strong level of performance on the unseen `hard' exploration games, despite no other mechanism of exploration being built into NAIT. Of lesser importance is the use of tricubic kernel which is shown to result in an increase in performance with only a small cost in computation speed in comparison to a uniform kernel (i.e. take the mean), however this increase is minor.

\subsection{Effect of Hyperparamters} \label{sec:sensitivity}
Higher representation dimension $F$ tends to be both slower, and perform worse as it is increased past $F=289$ (Figure \ref{fig:sweeps}). We hypothesise this is due to poorer recall of HNSW at higher dimensions overshadowing information preservation gained by a higher $F$. This could potentially be offset by increasing the recall of the lookup via the \texttt{ef} and \texttt{M} parameters, however this would come at the cost of a significant reduction in computational efficiency. Interestingly, as $F$ is held constant and \texttt{M} increased, performance is stable up to $M=160$, but drops sharply after this. This would indicate some level of noise in the value estimate is beneficial, and perfect recall is in fact not desirable for NAIT. Setting \texttt{M} too low however also degrades performance as expected. 

Such high performance with low $F$ indicates that the information required to learn basic play is present in a surprisingly small number of DCT coefficients. One implication for ATARI100k is that learning a good representation is significantly less important in early learning than being able quickly assign credit.

From our sweeps, $k$ appears to have a clear optimal value across all games (see fig \ref{fig:k_sweep}). This is expected as $k$ effectively balances local generalization with the smoothness of the value estimate, and setting $k$ too large removes information from the value estimate. In contrast, the choice of learning rate does not effect performance smoothly, however the high-level metrics are surprisingly strong even at large $\alpha$, displaying a level of robustness to the choice of $\alpha$ that departs significantly from neural-based approaches. One explanation is due to $\alpha$ not being considered at new $(s,a)$ pairs where the new value is set directly. This forces NAIT into a high-learning rate setting by default.

\section{Conclusion}
We have introduced a locally generalising approach to RL that is both sample and computationally efficient, achieving state of the art performance in the low sample regime on ATARI while requiring two orders of magnitude less walltime. Due to the very low walltime and computational requirements, we hope that future work adopts our algorithm as a benchmark when demonstrating the efficacy of learned model or learned representation approaches. Further work may investigate the possibility of our approach as a quick but simple initial learner that can merge into, or inform, a more complex learner that does produce useful representations, via mechanisms such as behavioural cloning or similar. This has exciting potential to significantly improve the sample efficiency of established methods while maintaining their asymptotic performance.

\clearpage
\bibliography{library}

\appendix
\clearpage

\section{Implementation Details}
\subsection{ATARI}
\begin{figure}
	    \centering
		\includegraphics[width=\columnwidth]{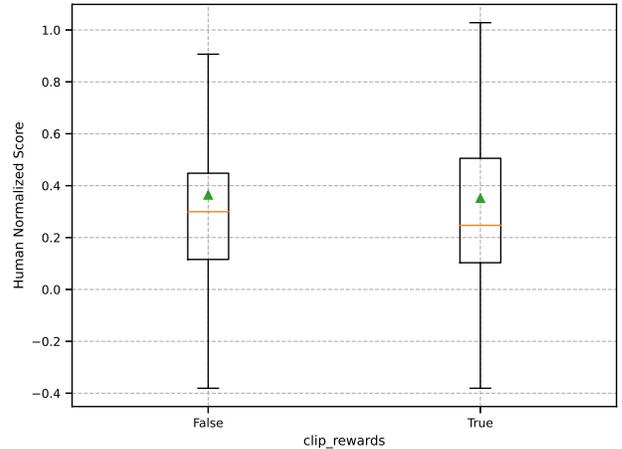}
	\caption{Effect of reward clipping on NAIT on ATARI100k.}
			\label{fig:cliprewards}
\end{figure}

\begin{figure}
	    \centering
		\includegraphics[width=\columnwidth]{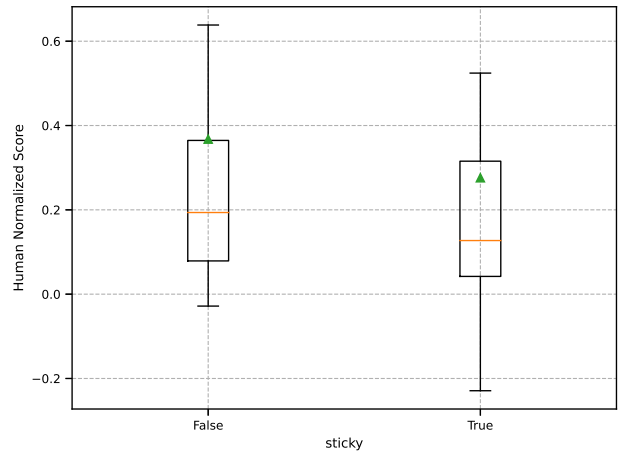}
	\caption{Effect of sticky actions on NAIT's performance over the entire 57 game set. Stochastic max 30 no-ops enabled for both True and False.}
			\label{fig:stickyactions}
\end{figure}

Throughout the literature it is common to observe slight alterations in the underlying environments used to benchmark approaches. This makes direct comparison difficult, and indeed among the current state-of the art approaches on the 100k domain the specifics of the ATARI setup do vary. We endeavored to choose parameters most similar to those recommended by the original authors, and not those that result in the maximum the raw score. Where our choices deviate (to facilitate comparison to previous methods), we run separate ablation studies with the recommended values (Fig. \ref{fig:cliprewards} and \ref{fig:stickyactions}). NAIT is robust to changes in the underlying environments, with reward clipping having little effect and sticky actions resulting in a minor decrease in performance, which is expected as it makes control more challenging. Notably the change in HNS distribution between the 57 and 26 game sets is minor, indicating that NAIT has not been overfitted to the 26 game set. 

\begin{table}
	\begin{center}
		\begin{tabular}{lrr}
			\toprule
			Hyperparameter & Symbol&  Value\\
			\midrule
			Number of Neighbours & $k$ & 64 \\
			State Dimension & $F$ & 289\\
			Discount & $\gamma$ & 0.99 \\
			Learning Rate & $\alpha$ & 0.1 \\
			\midrule
			HNSW settings\\
			\midrule
			\texttt{efc} && 200\\
			\texttt{efs} && 200\\
			\texttt{M} && 40\\
			Space && $\normltwo$\\
			\bottomrule
		\end{tabular}
	\end{center}
	\caption{Default hyperparameters for NAIT.}
	\label{app:hyperparams}
\end{table}

\begin{table*}
	\begin{center}
		\begin{tabular}{lrrrrr}
			\toprule
			Hyperparameter & Ours & D.E. Rainbow & SimPLe & SPR & Recommended\\
			\midrule
			Greyscale				& True & True   & True & True  & True  \\
			Observation size 		& (84, 84) & (84, 84) & (84, 84) & (84, 84) & (84, 84)\\
			Interpolation 			& Inter-linear & Inter-linear & Inter-area & Inter-area & n.a. \\
			Sticky Actions          & False & False & False & False & True \\
			Frameskip/Action Repeat & 4 & 4 & 4 &  4& n.a.\\
			Framepooling 			& True&  True & True & True & n.a. \\
			Colour Averaging		 & False  & False & False & False & n.a.\\
			Framestack 				& 4 & 4 & 4 & 4 & n.a.\\
			Terminal on loss of life & False & True & False & False & False \\
			Reward Clipping			 & False & True   & True & True & False \\ 
			Max No-Op Starts		& 30 & 30 & n.a. & 30 &  0\\
			Seeds 					& 10 & 5 & 10 & 10 & n.a.\\
			Max frames 				& 400k & 400k  & 400k  & 400k & n.a. \\
			\bottomrule
		\end{tabular}
	\end{center}
	\vskip -0.1in
	\caption{ATARI100k evaluation settings for various approaches. Recommended is from Machado \textit{et. al. }2018.}
	\label{app:atari}
\end{table*}

\subsection{NAIT}

Exact match lookups are computed efficiently via a hashtable. For fast hashing, we make use of the Extremely Fast Non-Cryptographic Hash library \texttt{xxhash}, available at \url{https://github.com/Cyan4973/xxHash}. To keep the memory overhead low, we initialize all buffers with a fixed capacity of 1000 samples, and double their capacity when filled.   
Because no representation learning is carried out, our approach does not require a significant number of hyper-parameters to be tuned. The final values are listed in Table \ref{app:hyperparams}. Note that HNSW parameters do noticeably effect performance, as they determine the recall of the approximate neighbour lookup.

\subsection{HNSW} \label{app:hnsw}
HNSW was chosen as the approximate nearest neighbour library as presents the best speed/recall performance on dense data at the time of publication. There are several implementations available; we make use of v0.5.0 of the header-only library available at \url{https://github.com/nmslib/hnswlib}. While there are some feature restrictions, this implementation supports incremental index construction (i.e. single data points can be added without requiring re-indexing), which is critical to our approach. HNSW's speed/recall tradeoff can be can be significantly altered via three parameters; \texttt{efs}, \texttt{efc} and \texttt{M}. \texttt{efs} is the size of the candidate list to keep during search - in general larger \texttt{efs} leads to higher recall but slower lookups. \texttt{efc} is the same parameter as \texttt{efs}, but refers to the value during \textit{construction} time, when a new point is being inserted. \texttt{M} determines the number of bi-directional links made for each node when constructing the proximity graph. Higher values also correspond to higher recalls at the cost of speed, and \texttt{M} also directly effects the memory requirements of the index.  

\begin{figure*}
	\begin{center}
		\centerline{\includegraphics[width=\textwidth]{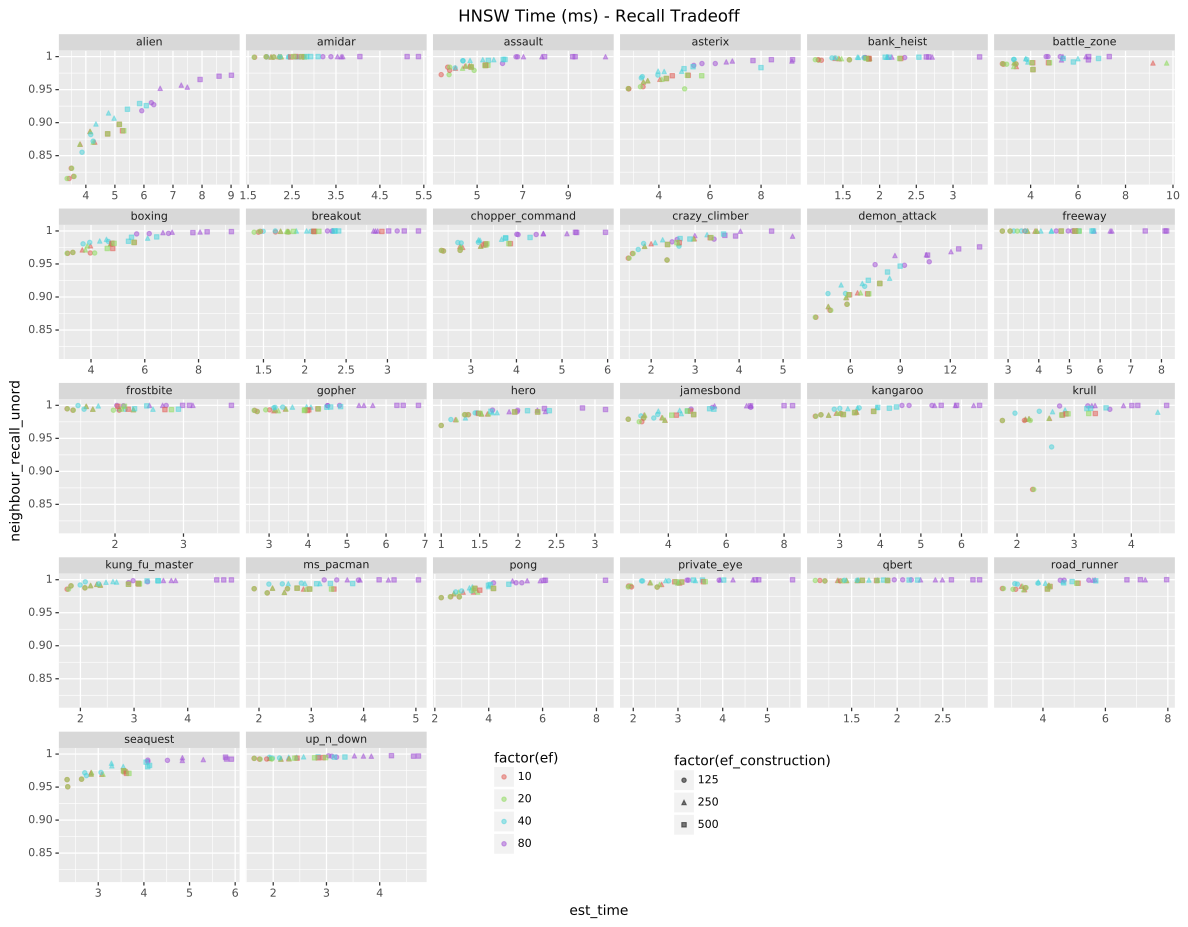}}
		\caption{Runtime/Recall tradeoff of HNSW parameters \texttt{ef} and \texttt{efc} for each game. Up and to the left is best.}
		\label{fig:hnswtradeoff}
	\end{center}
\end{figure*}

To quantify the effect of \texttt{efs} and \texttt{efc} on performance, we sampled 10000 frames from each game (with a random policy), applied the DCT transform with $F=225$ and calculated the recall for a query of $k=30$ neighbours for 1000 samples, over a range of values, averaging the result. Fig. \ref{fig:hnswtradeoff} displays the outcome. In general recalls are surprisingly high, even for low values of \texttt{efs} and \texttt{efc}, however the general trend is as expected, with higher values leading to increases in recall, but decreases in speed. Due to the high recall present in all runs, it is clear that inaccuracies in the value function due to poor lookup results will be minor for most games.

\begin{figure}	
	\centering
		\includegraphics[width=\columnwidth]{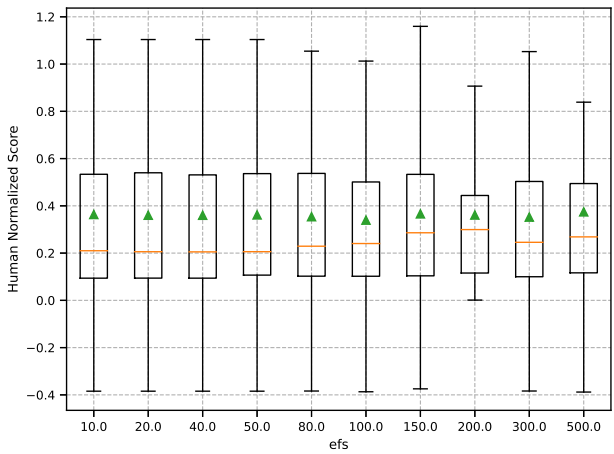}
	\caption{HNS distribution fine-grained sweep of HSNW parameter \texttt{efs} on ATARI100k}
	\label{fig:efs}
\end{figure}

\begin{figure}	
	\centering
		\includegraphics[width=\columnwidth]{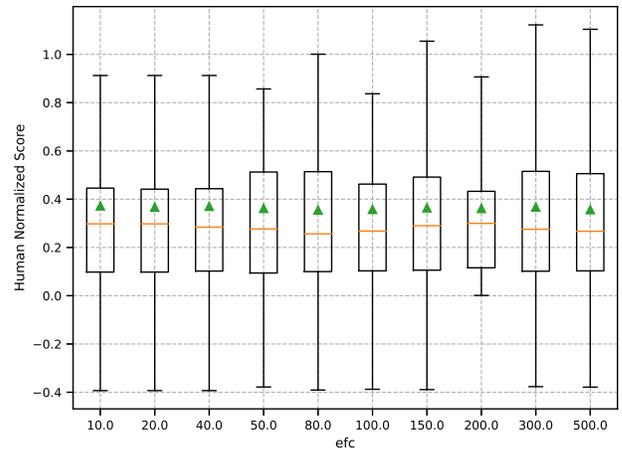}
	\caption{HNS distribution fine-grained sweep of HSNW parameter \texttt{efc} on ATARI100k}
	\label{fig:efc}
\end{figure}

In addition to quantifying the recall on random policy frames, we also carried out full benchmark sweeps over the \texttt{efs} and \texttt{efc} parameters. Higher \texttt{efs} tends to increase performance which is expected, however it does appear to have an optimal point. We hypothesise that a small error in recall may actually be desirable as it provides a mechanism for exploration, which we do not have explicitly builtin except in the case of a tiebreak. As the number of samples grows, the effect of poorer recall on value estimates is reduced (as an area becomes more densely covered), leading to a scheme similar to e-greedy on a decaying schedule. 

\begin{figure}	
	\centering
		\includegraphics[width=\columnwidth]{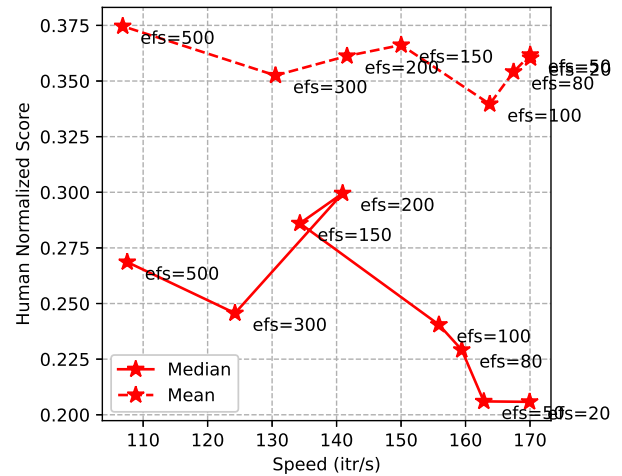}
	\caption{Speed-Performance curve of \texttt{efs} on ATARI100k.}
	\label{fig:discount1}
\end{figure}

\begin{figure}	
	\centering
		\includegraphics[width=\columnwidth]{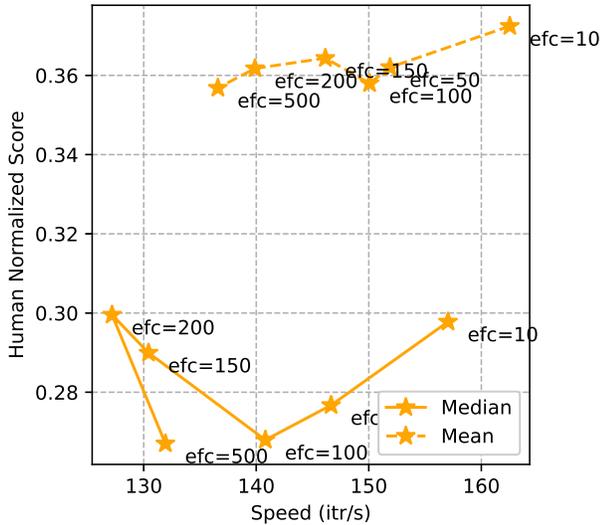}
	\caption{Speed-Performance curve of \texttt{efc} on ATARI100k.}
	\label{fig:discount2}
\end{figure}

\section{Further Analysis: Walltime}
The benefit of lower computational complexity is limited in many applications if it comes with a proportional decrease in sample efficiency. In ATARI for example, if NAIT runs twice as fast, but takes double the number of samples to meet the performance of competing methods, it's benefit is questionable. Here we make the comparison as explicit as possible and plot performance against walltime for a single representative game, `Ms Pacman' (`Ms Pacman' was chosen as it is often the `median game' in our experiments). We baseline against SPR - the current state of the art for sample efficiency. Note that NAIT runs in a single thread, while SPR requires multiple threads and access to a P100 GPU. We perform 10 separate runs and plot them on top of each other, performing no averaging or interpolation, logging the trailing 5-episode reward every 1000 interactions. We run NAIT for 250k steps. It takes NAIT $8$ minutes to achieve the performance of SPR@100k samples (which takes SPR 4.6 hours). 

\begin{figure}	
	\centering
		\includegraphics[width=\columnwidth]{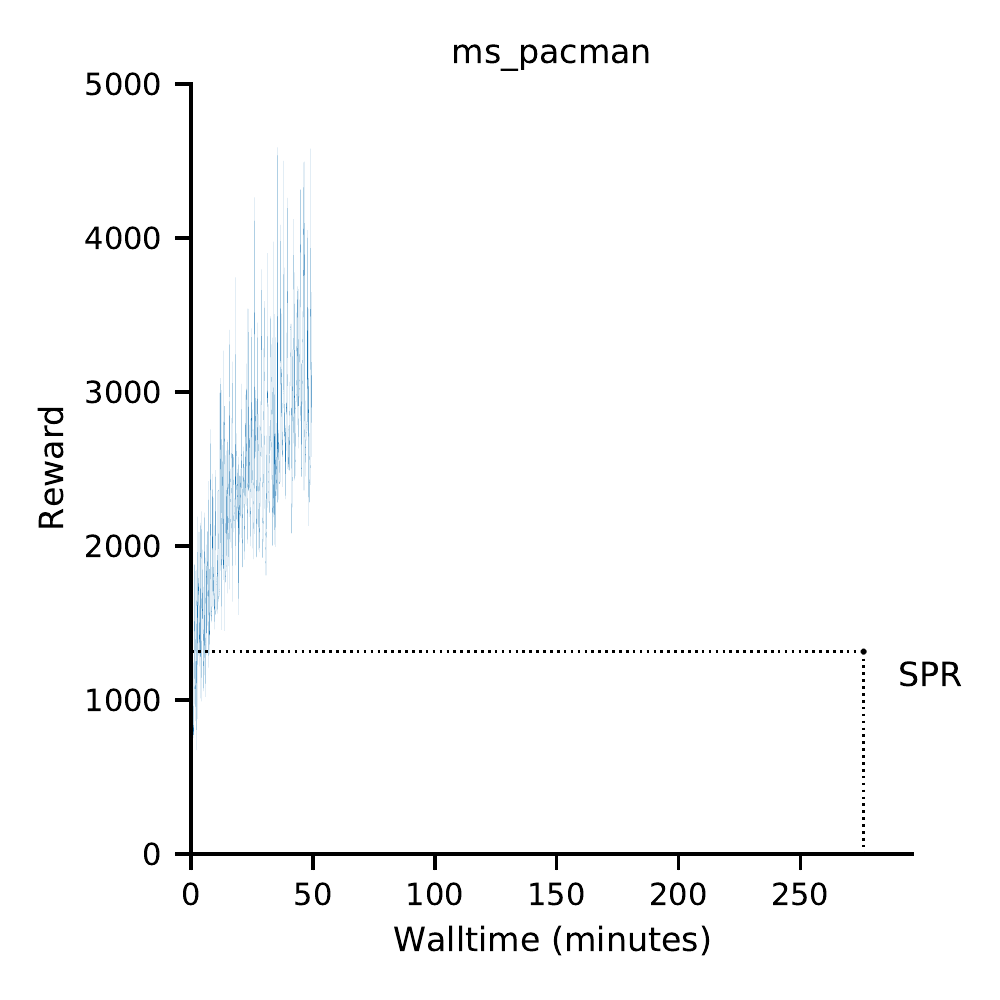}
	\caption{Raw score as a function of walltime for `median' game \texttt{Ms Pacman}.}
	\label{fig:discount3}
\end{figure}

\section{Further Discussion}
\subsection{Importance of Adaptive Kernel Interpolation} \label{app:interpolation}
\begin{figure*}
	\centerline{\includegraphics[width=1.\textwidth]{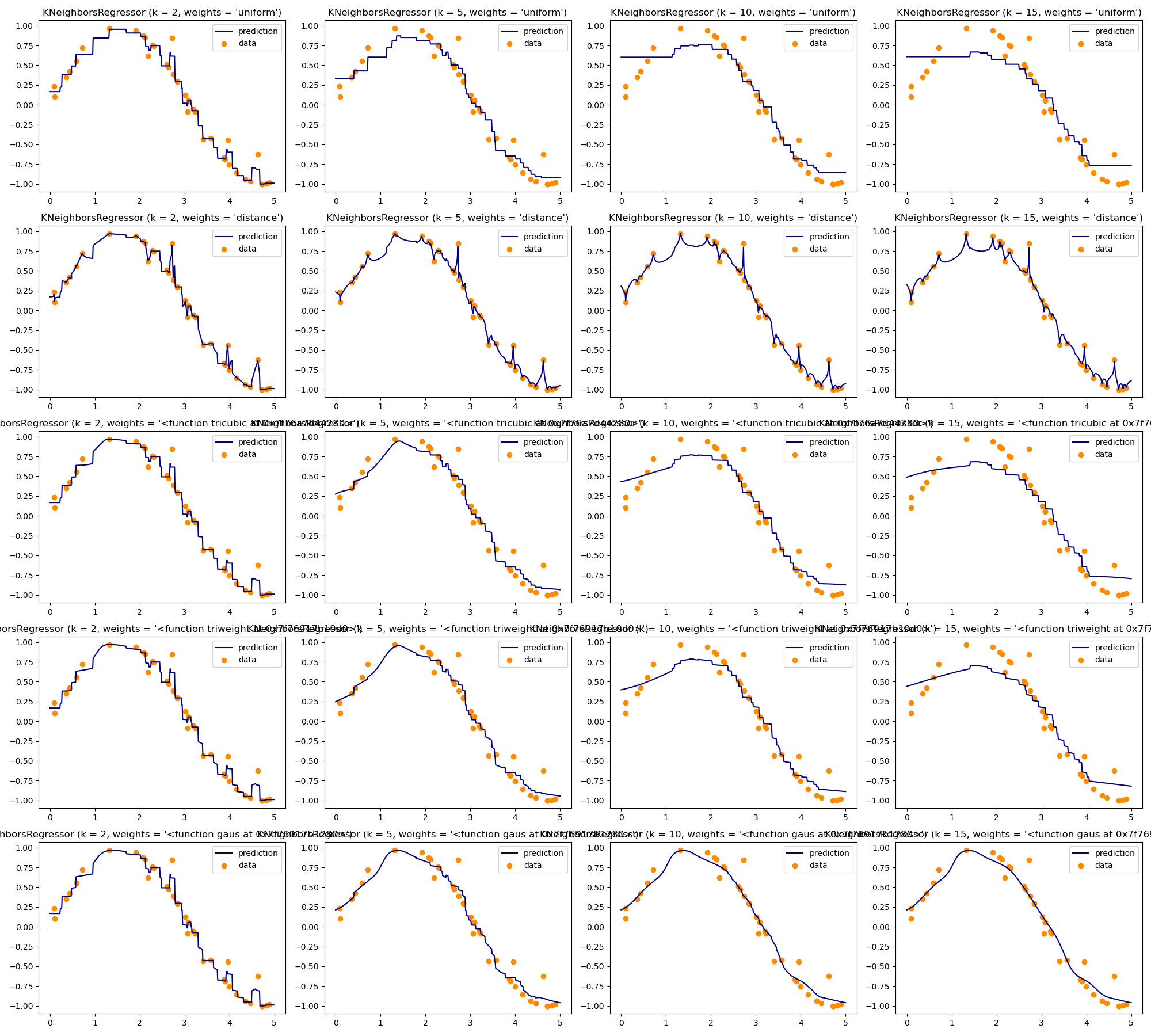}}
	\caption{Visualization of kernel type and $k$ on 1D interpolation quality. Rows represent a uniform, inverse distance, Tricubic, Triweight, and Gaussian kernel respectively, with columns indicating increasing $k$. Triweight and Tricubic can be seen to be highly dependent on $k$ for their quality, but provide broad generalization when the data is sparse (see the peaks in rows 3 and 4 of column 2), and behave similarly to a mean when the data is dense.}
	\label{fig:kernel}
\end{figure*}

The use of an adaptive kernel (i.e. one which does not require a fixed bandwidth hyperparameter) is important as the absolute distances between state representations will vary from game to game (as a function of how much pixel intensity varies). In addition, the type of smoothing applied by the kernel can have a significant effect on the final Q-value approximation. Inverse distance weighting, for example, which has been used in other work, creates a distinct ``spiky appearance'' which becomes more pronounced in higher dimensions as the space between samples grows. Under this interpolation very sharp changes in the value function can occur as one approaches an already sampled point, which is undesirable. Gaussian kernels in general produce the most appealing-looking estimates, but come with the aforementioned drawback of requiring a fixed bandwidth set at the start of training for all games. Normalized tricubic kernels provide a good compromise - however as can be seen their performance is highly dependent on $k$, hence the importance of its value in our overall algorithm. For emphasis, theses effects are visualized in Fig. \ref{fig:kernel} on sample 1D data.

\begin{figure*}
	\begin{center}
		\includegraphics[width=1.7\columnwidth]{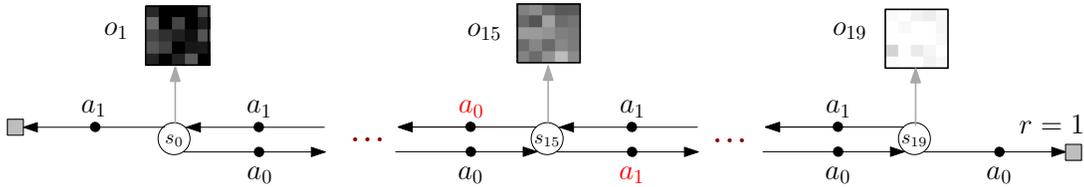}
		\caption{Transition diagram for the 19-state image random walk MDP. No reward is received except after taking $a_1$ in State 19. State 15 is a special state where the direction of actions is reversed. All transitions are deterministic. $\gamma=0.9$.}
		\label{fig:mdp}
	\end{center}
\end{figure*}

\subsection{Benefits of a DCT over Random Projection}
One surprising result in our work is the out-performance of random projection-based representations by a DCT-based representation. We hypothesize two main reasons for this;
\begin{enumerate}
	\item \textbf{Bias towards global representation} because we select DCT components in decreasing order of frequency, we bias towards large, global representations of the current game state. Tuning $F$ in this scenario modifies the level of granularity that is preserved in the representation. Low $F$ captures only broad gradients and shapes, while high $F$ captures greater detail but comes with the discussed downstream costs. By tuning $F$ we allow the algorithm to capture sufficient low-level detail (such as the location of the ball in pong) to facilitate basic play, but nothing more.
	\item \textbf{DCT coordinates are normalized.} Because DCT coordinates are already normalized no dimension is weighted higher than any other in the $k$-NN lookup. In contrast, if a random projection is used, one of the random bases may by chance cover pixels that have much higher average intensities than another. In this case, it will be weighted higher in the subsequent lookup as that dimension is ``stretched'' proportional to the intensity increase. This is a common problem when using $k$-NN lookups and is remedied by normalizing all dimensions to the same range. This is not possible in our case as the ranges of each dimensions are not known before learning begins. In this scenario and \textit{adaptive} normalization scheme, that tracks running statistics over each dimension and normalizes accordingly, may be necessary. 
\end{enumerate}
In regards to compute efficiency, the DCT is also faster than a standard projection. This is due to the highly optimized underlying fft libraries used to compute the DCT (\texttt{scipy.fftpack}), which run in $\mathcal{O}(DlogD)$, vs $\mathcal{O}(DF)$ for the random projection.

\subsection{Why Does a Nonparametric Approach Perform so Well on ATARI100K?}
A major advantage to ATARI100k is that because samples are limited, no forced forgetting is required to ensure memory limits are not exceeded, as the entire game history is less than the size of a standard replay buffer. However, given that our representation is quite naive, a better question is perhaps why value-function methods perform so poorly. We discuss several possibilities below. 

Value function approaches to Deep-RL learn a Q-function 
\begin{equation}
Q^\pi_\theta(s, a) = \mathbb{E}_\pi \left[G_t | s_t=s, a_t = a \right]
\end{equation}
from which a policy is derived, most commonly $\epsilon$-greedy~\cite{DQN}, although transformations via softmax and related operators are also used~\cite{softmax}. The Q-function is parameterised (indicated via subscript) by a DNN. Initial attempts~\cite{neuralq} to combine neural function approximation with Q-learning minimised the squared one-step TD error calculated on stored samples, 
\begin{equation}
\mathcal{L_{\text{DQL}}} = \left(Q^\pi_\theta(s, a) - \left(r_t + \gamma \max_a Q^\pi_\theta(s', a')\right)\right)^2.
\end{equation}
While this appears a reasonable generalisation of tabular Q-learning to high-dim observations, it is unstable and in most cases will not converge. This is due to the combination of function approximation, bootstrapping, and off policy training commonly known as the \textit{deadly triad}~\cite{deadlytriad}. Bootstrapping and function approximation are particularly damaging when the approximation is poor, as errors are immediately propagated, requiring low learning rates to avoid divergence. \textit{This limitation is highly pronounced in the low-sample domain, where any parameterization will be poor if initialized randomly.}

Another explanation is the method by which samples are incorporated. In general, for sample efficient learning we want to store and exploit raw information about the environment to the greatest extent possible as interactions are the constraining factor. Experience replay does store raw experiences (although if n-step returns are used they become policy dependent), but traditional approaches do not fully exploit this information. This is due to the under-use of available data due to limited training steps performed on the value network combined with the necessity of low learning rates. As a consequence, all available information is not fully incorporated into the Q-function approximation early during training. The original DQN for example, uses 4 interactions with the environment for each training step. SPR, in contrast, uses the highest published ratio of 2 training steps per interaction with the environment, and in ablation studies showed increasing this parameter significantly improves sample efficiency. However even if all samples in the buffer are sampled at some point for training (which is unlikely), due to the low learning rate their information will still not  be fully incorporated into the Q-function parameterization.

\subsection{Effect of the Discount Rate}
\begin{figure}	
		\includegraphics[width=\columnwidth]{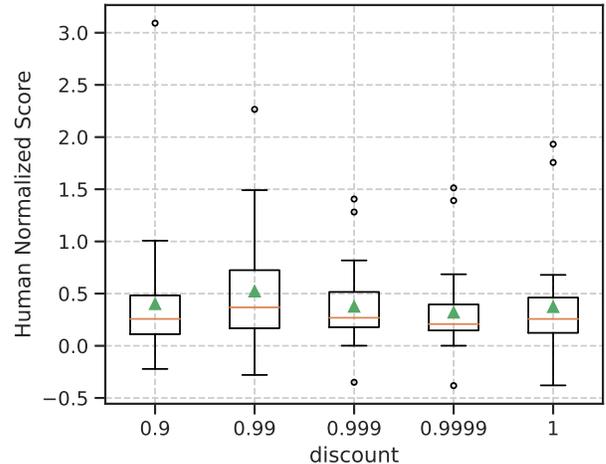}
	\caption{Effect of $\gamma$ on total HNS on ATARI100k}
	\label{fig:discount}
\end{figure}

The actual objective being optimised is total reward, and so to maximise directly what we care about, $\gamma=1$ would seem appropriate. As others have noted, however \cite{GAE}, often a surrogate objective created by a lower discount rate can produce better results. We observe this phenomenon here, with the optimal discount found to be 0.99, lower than is traditional. This is likely due to the much shorter trace lengths that occur when the interaction budget is limited (as the agent does not reach superhuman levels of play that result in large game lengths). 

Figure \ref{fig:discount} shows the effect of various discount rates on HNS over time. Interestingly, we observe for some games a clear demarcation in  performance induced by a varied $\gamma$. This is most pronounced in the game \texttt{freeway}, where higher discounting significantly improves final performance. For freeway, this is likely caused as low discounting results in the agent repeating the first route it finds to the ``goal'' (crossing the street), regardless of its efficiency. More aggressive discounting directly pushes the agent towards short paths, which allows it to reach a higher final score within the time-limit.

\begin{table*}
	\label{app:numerical_results}
		\centering
		\begin{small}
			\begin{sc}

\begin{tabular}{@{}lrrrrrrr@{}}
\toprule
\textbf{Game}         & \textbf{Random}      & \textbf{Human}                & \textbf{100K} & \textbf{250k} & \textbf{500k} & \textbf{750k} & \textbf{1M} \\ \midrule
Alien*                & 227.8                & 7127.7                        & 918.35        & 1072.5        & 1433.3        & 1836.8        & 2057.2      \\
Amidar*               & 5.8                  & 1719.5                        & 213.37        & 377.065       & 481.875       & 592.225       & 715.96      \\
Assault*              & 222.4                & 742                           & 483.21        & 509.145       & 524.685       & 537.495       & 537.285     \\
Asterix*              & 210                  & 8503.3                        & 900.5         & 1336.25       & 1920.75       & 2678          & 3294.5      \\
Asteroids             & 719.1                & 47388.7                       & 606.15        & 982.55        & 1134.75       & 1261.7        & 1431.55     \\
Atlantis              & 12850                & 29028.1                       & 32268.5       & 45540.5       & 57116.5       & 79494         & 83677.5     \\
Bank Heist*           & 14.2                 & 753.1                         & 98.3          & 143           & 176.25        & 197.2         & 204.55      \\
Battle Zone*          & 2360                 & 37187.5                       & 9595          & 9930          & 12110         & 14250         & 14675       \\
Beam Rider            & 363.9                & 16826.5                       & 622.7         & 655.48        & 764.86        & 856.54        & 857.42      \\
Berzerk               & 123.7                & 2630.4                        & 564.2         & 789.1         & 882.15        & 893.95        & 911.1       \\
Bowling               & 23.1                 & 160.7                         & 74.01         & 86.435        & 86.325        & 90.47         & 91.09       \\
Boxing*               & 0.1                  & 12.1                          & 5.135         & 10.97         & 15.44         & 18.57         & 24.4        \\
Breakout*             & 1.7                  & 30.5                          & 11.665        & 18.865        & 40.815        & 37.6          & 48.025      \\
Centipede             & 2090.9               & 12017.1                       & 5369.32       & 7076.52       & 10295.98      & 11805.91      & 12440.17    \\
Chopper Command*      & 811                  & 7387.8                        & 1711          & 2368          & 2592          & 4158          & 3893        \\
Crazy Climber*        & 10780.5              & 35829.4                       & 1240.5        & 3147          & 6092          & 7908          & 8506.5      \\
Defender              & 2874.5               & 18688.9                       & 4451          & 6540.5        & 8706.5        & 9700.75       & 11655.5     \\
Demon Attack*         & 152.1                & 1971                          & 262.025       & 332           & 389.275       & 458.175       & 458.175     \\
Double Dunk           & -18.6                & -16.4                         & -5.39         & -6.6          & -3.07         & -6.63         & -5.99       \\
Enduro                & 0                    & 860.5                         & 4.345         & 4.715         & 8.95          & 12.325        & 17.44       \\
Fishing Derby         & -91.7                & -38.7                         & -80.92        & -74.035       & -68.575       & -62.67        & -58.325     \\
Freeway*              & 0                    & 29.6                          & 25.09         & 25.865        & 26.1          & 26.04         & 26.075      \\
Frostbite*            & 65.2                 & 4334.7                        & 1808.8        & 2783.85       & 3610.8        & 4029.3        & 4462.6      \\
Gopher*               & 257.6                & 2412.5                        & 979           & 1528.9        & 2597.5        & 2701.9        & 4315.7      \\
Gravitar              & 173                  & 3351.4                        & 451           & 594.75        & 727.75        & 828           & 909         \\
Hero*                 & 1027                 & 30826.4                       & 5311.2        & 5524.25       & 5538.225      & 5573.275      & 5576.875    \\
Ice Hockey            & -11.2                & 0.9                           & -8.965        & -7.97         & -5.91         & -5.08         & -4.65       \\
Jamesbond*            & 29                   & 302.8                         & 306.25        & 418.5         & 482.25        & 499.5         & 496.75      \\
Kangaroo*             & 52                   & 3035                          & 983           & 1229          & 1818.5        & 1915          & 2038.5      \\
Krull*                & 1598                 & 2665.5                        & 3588.64       & 3851.905      & 4227.73       & 4570.51       & 4733.92     \\
Kung Fu Master*       & 258.5                & 22736.3                       & 10408         & 17178         & 24448         & 26809.5       & 30329.5     \\
Montezuma Revenge     & 0                    & 4753.3                        & 0             & 0             & 0             & 35            & 80          \\
Ms Pacman*            & 307.3                & 6951.6                        & 2579.2        & 2951.15       & 3412.45       & 3654.1        & 3764.7      \\
Name This Game        & 2292.3               & 8049                          & 3522.05       & 4196.75       & 5241.4        & 5173.55       & 5914.9      \\
Phoenix               & 761.4                & 7242.6                        & 1448.3        & 2467.9        & 3729.45       & 4563.8        & 4852.5      \\
Pitfall               & -229.4               & 6463.7                        & -166.855      & -29.79        & -198.64       & -77.075       & -3.015      \\
Pong*                 & -20.7                & 14.6                          & -7.89         & 2.76          & 12.27         & 14.6          & 15.225      \\
Private Eye*          & 24.9                 & 69571.3                       & 105.38        & -153.495      & -209.215      & -88.195       & -54.48      \\
Qbert*                & 163.9                & 13455                         & 2361.125      & 4475.125      & 6273.75       & 8291.375      & 9557        \\
Riverraid             & 1338.5               & 17118                         & 2634.95       & 3615.65       & 4409.35       & 5061.2        & 5247.3      \\
Road Runner*          & 11.5                 & 7845                          & 7113.5        & 12953.5       & 14113         & 16166         & 17408       \\
Robotank              & 2.2                  & 11.9                          & 4.905         & 4.89          & 5.845         & 5.425         & 6.395       \\
Seaquest*             & 68.4                 & 42054.7                       & 417.1         & 596.9         & 778.2         & 830.4         & 927.7       \\
Skiing                & -17098.1             & -4336.9                       & -30372.155    & -30623.19     & -30664.61     & -30901.3      & -30696.87   \\
Solaris               & 1236.3               & 12326.7                       & 967.3         & 1381.9        & 1788.4        & 1750.2        & 2622.5      \\
Space Invaders        & 148                  & 1668.7                        & 646.125       & 999.225       & 1289.825      & 1304.5        & 1428.325    \\
Star Gunner           & 664                  & 10250                         & 1324.5        & 1401          & 1400          & 1386.5        & 1347        \\
Tennis                & -23.8                & -8.3                          & -4.89         & -5.855        & -6.96         & -5.735        & -5.605      \\
Time Pilot            & 3568                 & 5229.2                        & 2821.5        & 3284.5        & 4161.5        & 5159          & 5646.5      \\
Tutankham             & 11.4                 & 167.6                         & 99.68         & 111.31        & 104.535       & 78.54         & 81.63       \\
Up N Down*            & 533.4                & 11693.2                       & 7657.75       & 12082.95      & 18006.6       & 22985.4       & 27364.05    \\
Venture               & 0                    & 1187.5                        & 129           & 283.5         & 495           & 481.5         & 582.5       \\
Video Pinball         & 0                    & 17667.9                       & 5901.24       & 7280.085      & 9682.72       & 12524.755     & 12422.07    \\
Wizard OfWor          & 563.5                & 4756.5                        & 1041          & 1353          & 3337          & 4234          & 6451.5      \\
Yars Revenge          & 3092.9               & 54576.9                       & 6015.505      & 7973.015      & 8920.34       & 11150.245     & 12841.105   \\
Zaxxon                & 32.5                 & 9173.3                        & 2818.5        & 4850.5        & 6339.5        & 7081          & 7864.5      \\ \midrule
ATARI100K Mean HNS    & \multicolumn{1}{l}{} & \multicolumn{1}{l}{\textbf{}} & 0.362         & 0.529         & 0.700         & 0.794         & 0.904       \\
ATARI 100k Median HNS & \multicolumn{1}{l}{} & \multicolumn{1}{l}{\textbf{}} & 0.312         & 0.395         & 0.467         & 0.606         & 0.606       \\
Mean HNS              & \multicolumn{1}{l}{} & \multicolumn{1}{l}{\textbf{}} & 0.365         & 0.488         & 0.665         & 0.733         & 0.828       \\
Median HNS            & \multicolumn{1}{l}{} & \multicolumn{1}{l}{\textbf{}} & 0.176         & 0.265         & 0.436         & 0.501         & 0.520       \\ \bottomrule
\end{tabular}
			\end{sc}
		\end{small}
	\caption{Full numerical results for NAIT. HNS is Human Normalized Score, equal to $\frac{ \text{agent} - \text{random} }{ \text{human} - \text{random} }$. For our results, at the completion of training, 50 episodes are run for 10 seeds over each game, with the final number averaged first over episodes, then over seeds. $^*$ Indicates the game belongs to the ATARI100k subset.} 
\end{table*}

\begin{figure*}
	\centerline{\includegraphics[width=\textwidth]{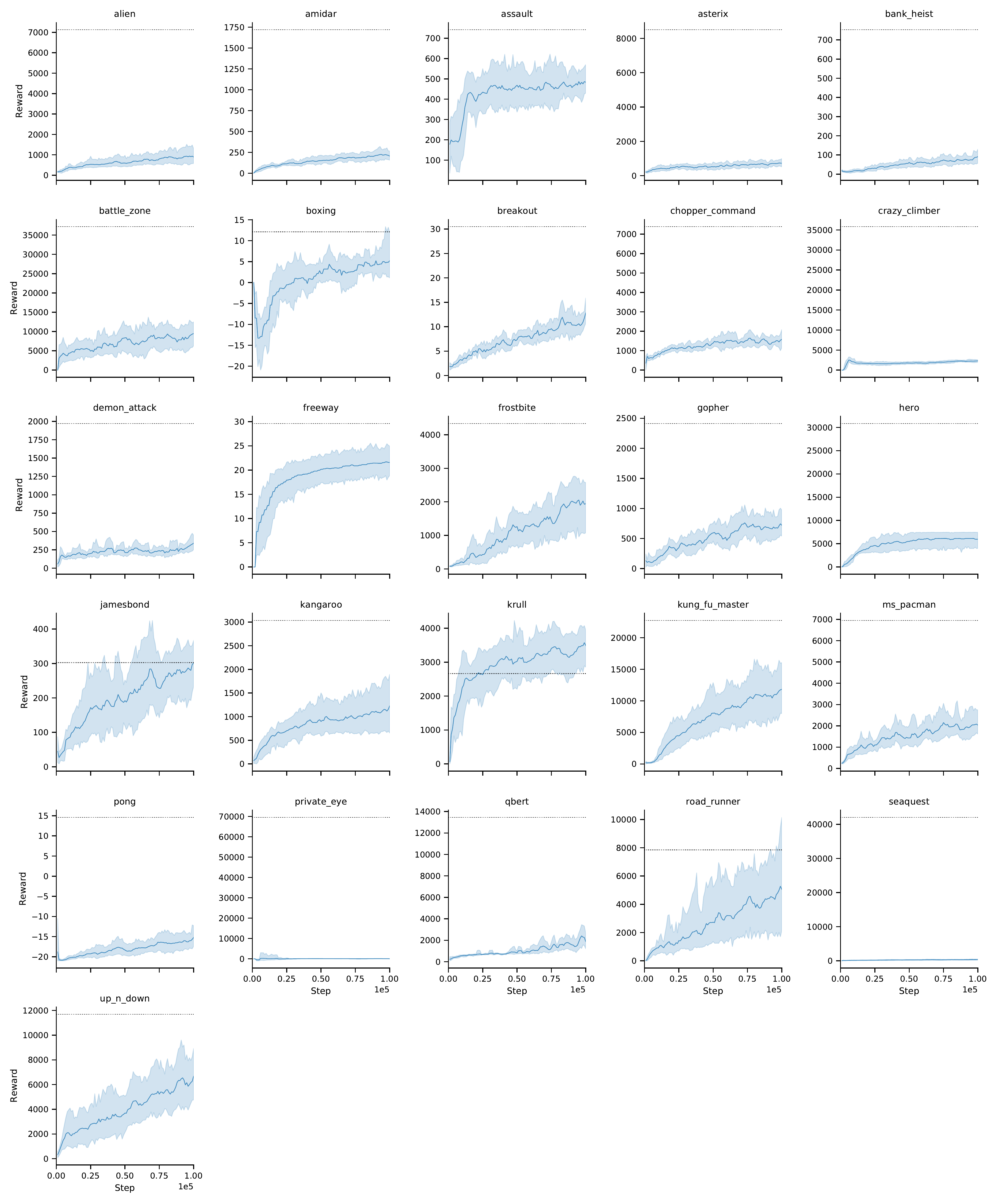}}
	\caption{Game-by-game performance over 100k interactions for the 26 game `ATARI100K' subset. Plots show trailing 5 episode average reward logged every 1000 interactions. Centre line is mean over 10 seeds, shaded area indicates the 90\% confidence interval. Human Expert Score is indicated by horizontal black line.} 
	\label{fig:full_runatari}
\end{figure*}

\begin{figure*}
	\centerline{\includegraphics[width=.9\textwidth]{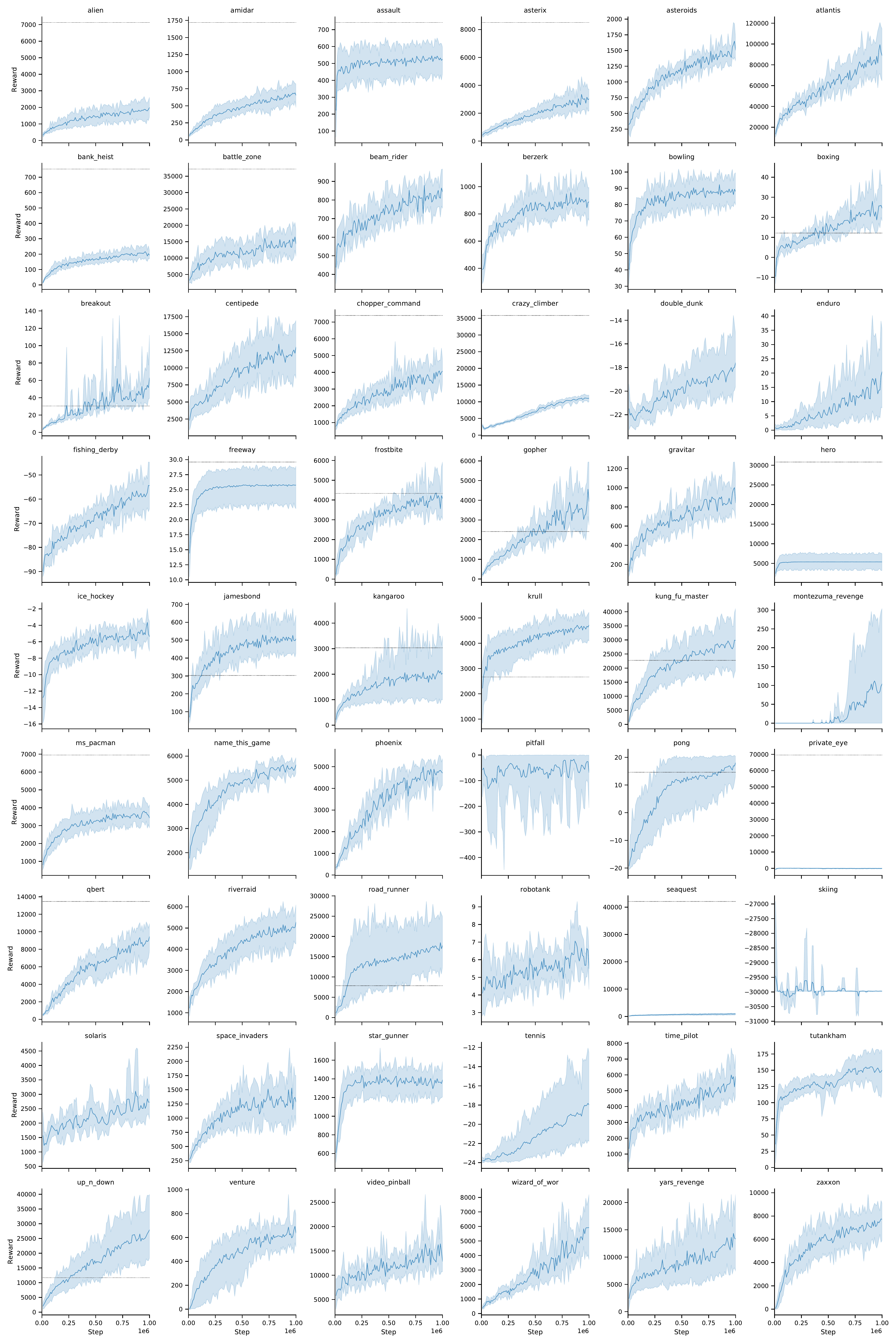}}
	\caption{Game-by-game performance over 1M interactions for all games with hyperparameters tuned for the 26-game subset at 100k. Plots show trailing 5 episode average reward logged every 5000 interactions. Centre line is mean over 10 seeds, shaded area indicates the 90\% confidence interval. Human Expert Score is indicated by horizontal black line on the 26 game subset.} 
	\label{fig:full_run}
\end{figure*}

\end{document}